%% file: main.tex
\PassOptionsToPackage{table}{xcolor}

\documentclass{article}
\usepackage[preprint]{icml2026}

\usepackage[utf8]{inputenc}    %
\usepackage[T1]{fontenc}       %

\usepackage{amsmath}
\usepackage{amssymb}
\usepackage{amsfonts}          %
\usepackage{amsthm}            %
\usepackage{nicefrac}          %

\usepackage{booktabs}          %
\usepackage{multirow}          %
\definecolor{rowshade}{gray}{0.93}

\usepackage{algorithm}
\usepackage{algpseudocode}

\usepackage{graphicx}          %
\usepackage{subcaption}        %
\usepackage{wrapfig}           %

\usepackage{xcolor}            %
\usepackage{microtype}         %
\usepackage{enumitem}          %
\usepackage{natbib}            %

\usepackage[disable,textsize=tiny]{todonotes}

\usepackage{url}               %
\usepackage{hyperref}          %

\usepackage{xurl}        %
\hypersetup{colorlinks=true, urlcolor=blue, linkcolor=blue, citecolor=blue}

\theoremstyle{plain}

\theoremstyle{definition}

\theoremstyle{remark}

\icmltitlerunning{Individual Parameters in Weight-Sparse Transformers Appear Interpretable}

\begin{document}

\twocolumn[
  \icmltitle{
  Individual Parameters in Weight-Sparse Transformers Appear Interpretable} 
  \icmlsetsymbol{equal}{*}

  \begin{icmlauthorlist}
    \icmlauthor{Arnau Marin-Llobet}{sch}
    \icmlauthor{Stefan Heimersheim}{}
  \end{icmlauthorlist}

  \icmlaffiliation{sch}{School of Engineering and Applied Sciences, Harvard University, Cambridge, MA 02138}

  \icmlcorrespondingauthor{Arnau Marin-Llobet}{amarinllobet@seas.harvard.edu}

  \icmlkeywords{LLM safety, LLM interpretability, two-sample test, anomaly detection}

  \vskip 0.3in
]

\printAffiliationsAndNotice{}  %

\input{sections/1_content.tex}
\input{sections/2_results.tex}
\input{sections/2b_casestudies}
\input{sections/3_conclusion.tex}
\section{Acknowledgements}
This work was funded by Pivotal Research. Arnau Marin-Llobet is supported by Coefficient Giving and the RCC-Harvard Fellowship. Thanks to Francisco Ferreira da Silva, Zheng (Zack) Hui, Reilly Haskins, Motahareh Sohrabi, Yonatan Belinkov and Demba Ba for helpful discussions.

\bibliography{example_paper}
\bibliographystyle{icml2026}

\newpage
\appendix
\onecolumn
\input{sections/4_appendix}

\end{document}

%% file: sections/1_content.tex
\begin{abstract}
A central goal of mechanistic interpretability is to understand how neural networks work and what each individual component does. Dominant circuit-finding approaches focus on a specific behavior and reverse-engineer the role of components on the associated sub-distribution. However, past work has shown that components can have different functions that are active on different subsets of the input distribution. In this work we ask whether a single weight can be understood globally across the full training distribution by characterizing when it matters (the inputs on which ablating it changes the model's predictions). We introduce an automated LLM pipeline that writes a short, human-readable description of when a weight matters and verifies it on held-out text, crediting a weight only if its description generalizes. Across two sparse and two dense transformers, the fraction of weights that are interpretable (in this sense) is higher in sparse transformers than in dense ones, a gap that widens once unreliable descriptions are discarded. Our results show that a meaningful fraction of a sparse transformer model's weights can be interpreted: 12 to 31\% of weights have a single short description that identifies what the weight is used for\footnote{Project website: \href{https://weightpedia.org/individual-parameters-in-sparse-transformers/}{weightpedia.org/individual-parameters-in-sparse-transformers/}.
}
\end{abstract}

\section{Introduction}
While large language models have rapidly increased in capability, we are still far from understanding how they work. Mechanistic interpretability aims to reverse-engineer neural networks into the algorithms they implement, and the dominant approach does so by isolating circuits responsible for specific behaviors \cite{ferrando2024primer, olah2020zoom, wang2022interpretability, olsson2022context}. A circuit tells us what a set of weights does in service of one task on one distribution, but the same weight can participate in many circuits, contributing differently to each~\cite{elhage2022superposition, jermyn2023polysemantic}.

\begin{figure*}[t]
    \centering
    \includegraphics[width=\textwidth]{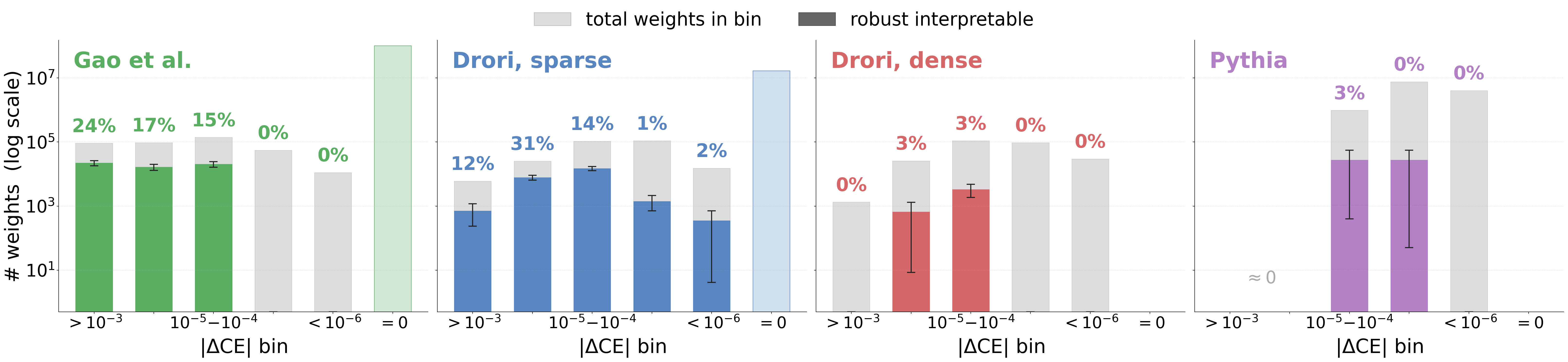}
    \caption{Per-weight causal impact distribution on a held-out distribution. Gray bars: total nonzero weights per $|\Delta\mathrm{CE}|$ bin. Colored bars within each gray bar show the fraction of weights in that bin that pass our interpretability test. The right-most column collects parameters that are exactly zero. Error bars are $\pm$1~SEM. {Pooled interpretability scores across all nonzero weights:} Gao et al.\ (2025) $15\%$, Drori (2026) sparse $9.6\%$, Drori (2026) dense $1.5\%$, Pythia-70m $0.4\%$.}
    \label{fig:ce_distribution}
\end{figure*}

As an alternative, we ask whether we can understand what an individual weight does on the full training distribution. In standard dense models, this question is difficult to investigate due to superposition: the weight basis is not an interpretable basis of the network's parameters \cite{gurnee2023finding, bricken2023monosemanticity, frankle2018lottery}, with rare exceptions \cite{yu2024superweight}. Recent work on weight-sparse transformers \citep{gao2025} addresses this by training models in which most weights are forced to zero, leaving a small number of surviving weights that organize into compact, readable circuits while the model retains some language modeling capability. 

This makes weight-sparse transformers a natural testbed for our question. The zero weights are trivially interpretable since they contribute nothing to the model's behavior; the open question is whether the surviving nonzero weights are interpretable too. These models let us develop weight-based interpretability pipelines today; once the field identifies an interpretable basis in parameter space for dense models (e.g.\ \cite{olah2025interference, braun2025interpretability}), we hope to apply these pipelines to frontier language models.

What would it mean to find that a weight is interpretable? We make a specific choice: we try to explain \emph{when} a weight matters; that is, the inputs on which it has a measurable effect on the model's output. This does not capture everything one might want from a full understanding of a weight, but it is concrete, testable, and a natural starting point. For each weight we propose a short, human-readable explanation of when it should be active, and we evaluate the explanation by checking whether it holds on data the explanation was not generated from.

Doing this at scale is hard for two related reasons. The first is \emph{automation}: a sparse transformer has tens of thousands of nonzero weights, and any useful test must run across all of them without human inspection of each one. The second is \emph{measurement}: a useful score must capture two properties at once. When the explanation claims a weight is active, ablating it should reduce model performance on those inputs; when the explanation claims the weight is inactive, ablating it should leave performance unchanged. Without the first, the explanation does not account for the weight's effect; without the second, the explanation passes trivially by being too broad.

In this paper we introduce an automated, LLM-based pipeline that addresses both challenges. For each nonzero weight we ablate it and measure the resulting Kullback--Leibler (KL) divergence and cross-entropy (CE) shift on a held-out corpus, prompt an LLM to generate candidate Python functions describing the token contexts most affected by the ablation, and score each candidate, taking the best of $n$ candidates as the explanation. We test the pipeline on a weight-sparse coding transformer \citep{gao2025}, a sparse-and-dense SimpleStories pair \citep{drori2026}, and Pythia-70m \citep{biderman2023} as a dense pretrained control. We find that the pipeline assigns interpretable predicates to up to $20$--$30\%$ of nonzero weights in the sparse models that have a high CE impact, compared to $0$--$3\%$ in the dense baselines when re-evaluated on a held-out corpus. Combined with the trivially interpretable zeros, this means a large fraction of parameters in weight-sparse transformers admit per-weight explanations, while the same is not true of dense models. Our contributions are as follows.

\begin{itemize}
\item \textbf{A framework for per-weight interpretability.} We define what it means for a single weight to be interpretable in terms of two complementary checks. \emph{Recovery} measures how much of the weight's ablation effect a candidate explanation accounts for. \emph{Inverse} measures the effect recovered by the explanation's negation, which guards against explanations that succeed only by being too broad.
\item \textbf{An automated LLM pipeline.} Given a weight, the pipeline profiles its ablation effect across a corpus and returns a scored explanation of when the weight matters, with no manual inspection required.
\item \textbf{Empirical evidence that a substantial fraction of nonzero parameters in weight-sparse transformers are individually interpretable}, far above what is achievable in dense controls, and the gap survives held-out validation.
\end{itemize}

\section{Related work}
\begin{figure*}[t]
    \centering
    \includegraphics[width=\textwidth]{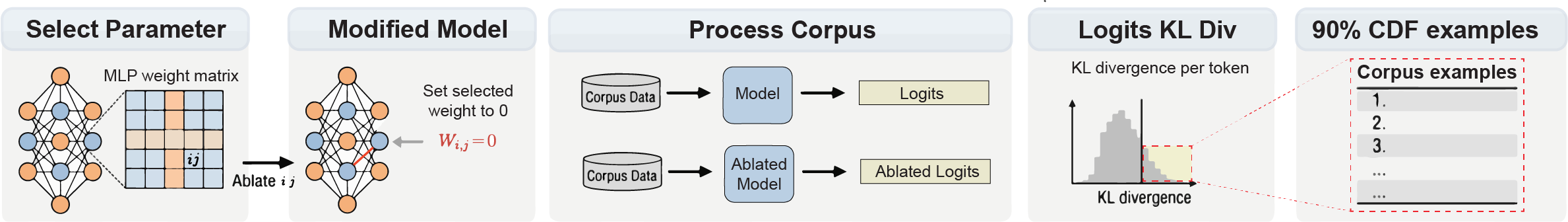}
    \caption{Per-weight interpretability pipeline. Given a target weight $W_{i,j}$ in an MLP layer, we run both the original model and an ablated copy on a corpus, measure the per-token KL divergence between them, and extract the token contexts that account for the top 90\% of cumulative KL. These contexts are passed to an LLM, which proposes candidate predicates describing them; the highest-scoring candidate becomes the weight's interpretation.}
    \label{fig:procedure}
\end{figure*}
\paragraph{Circuit decomposition from weight matrices.} A line of work going back to the original Circuits Thread \citep{olah2020zoom, elhage2021mathematical} and continued in recent bilinear-MLP analyses \citep{pearce2024bilinear} examines weight matrices to extract circuits responsible for specific behaviors. The unit of analysis is the circuit; weights enter as ingredients of larger structures rather than as objects of explanation. Approaches in this style identify a behavior, isolate the subnetwork that implements it, and read off the weight matrices that compose it. Recent work shows that this behavior-first localization extends even to broad capabilities in dense LLMs: harmful generation, for instance, depends on a compact, prunable set of weights \citep{orgad2026large}. All of these approaches are powerful when the target behavior is known in advance, but none produces a global account of what the model contains. Our approach inverts the dependency: we ask what each individual weight does without first committing to a behavior, which lets us characterize weights that participate in no named circuit at all, at a scale that hand-curated circuit extraction cannot reach.

\paragraph{Interpretable-by-design and sparse-weight transformers.} Several lines of work have produced transformers whose weights admit direct interpretation. Tracr \citep{lindner2023} compiles models from a high-level program (RASP; \citealp{weiss2021thinking}), so the interpretability of the resulting weights is a property of the compilation pipeline rather than something the model acquired from a training objective. A second line of work modifies the training recipe itself to encourage interpretability, by introducing an activation function (the Softmax Linear Unit) that increases the fraction of MLP neurons responding to a single human-interpretable concept without sacrificing language modeling performance \citep{elhage2022solu}. In our setting, \citet{gao2025} train transformers in which most weights are forced to zero, and show that the surviving weights organize into compact, readable circuits for specific behaviors. \citet{drori2026} extends this with the SimpleStories sparse-and-dense pair, which couples a sparse reasoning core to a dense output head and recovers more of the loss-capability frontier than fully sparse training. Our contribution is complementary; the sparse models are our testbeds \citep{gao2025, drori2026}. Where prior work establishes that sparse training produces interpretable circuits, we ask whether the individual parameters of these models admit per-weight explanations without first choosing a behavior.

\paragraph{Parameter decomposition.} A second line of work treats parameters as the unit of analysis but decomposes them into a learned basis of components rather than interpreting the raw entries. APD \citep{braun2025interpretability}, SPD \citep{bushnaq2025stochastic}, and L3D \citep{chrisman2025identifying} all follow this strategy. The motivation is that decomposed components may correspond to recognizable functions even when individual weights do not, which our results confirm is the case for dense models. We take a different approach. In sparse models, the raw weights are already a candidate basis for interpretation, and no decomposition step is needed. Working with the parameters as trained avoids the question of whether a learned decomposition reflects the model's computation or the optimization pressure of the decomposition objective. It also yields per-parameter rather than per-component answers, which is the relevant unit when the goal is to identify or edit specific parts of the model. 
\begin{figure*}[t]
    \centering
    \includegraphics[width=0.9\textwidth]{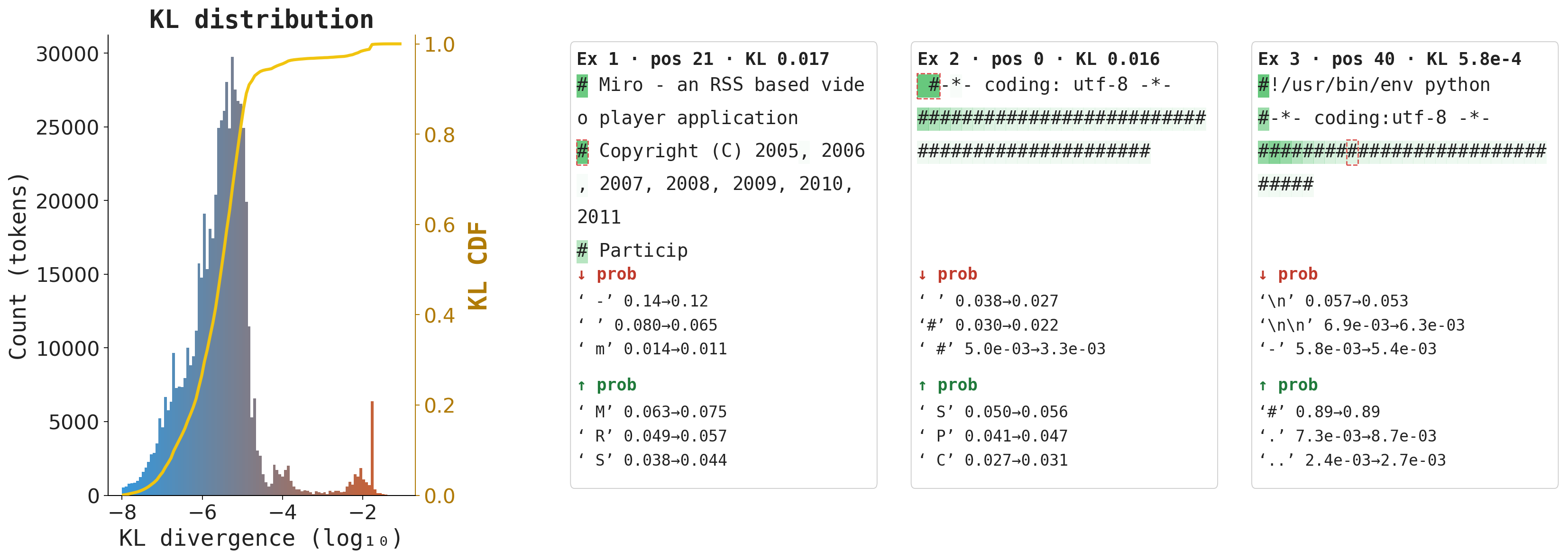}
    \caption{An interpretable weight from \citet{gao2025} sparse code transformer. Left: distribution of per-token KL divergence over the corpus when this weight is ablated, with the cumulative KL shown in yellow. Right: three token contexts drawn from the top of the KL distribution. Each panel shows the surrounding text, the position of the affected token, and the largest probability shifts (red: probabilities decreased by ablation; green: increased).}
    \label{fig:example_gao_a}
\end{figure*}

\begin{figure*}[t]
    \centering
    \includegraphics[width=0.9\textwidth]{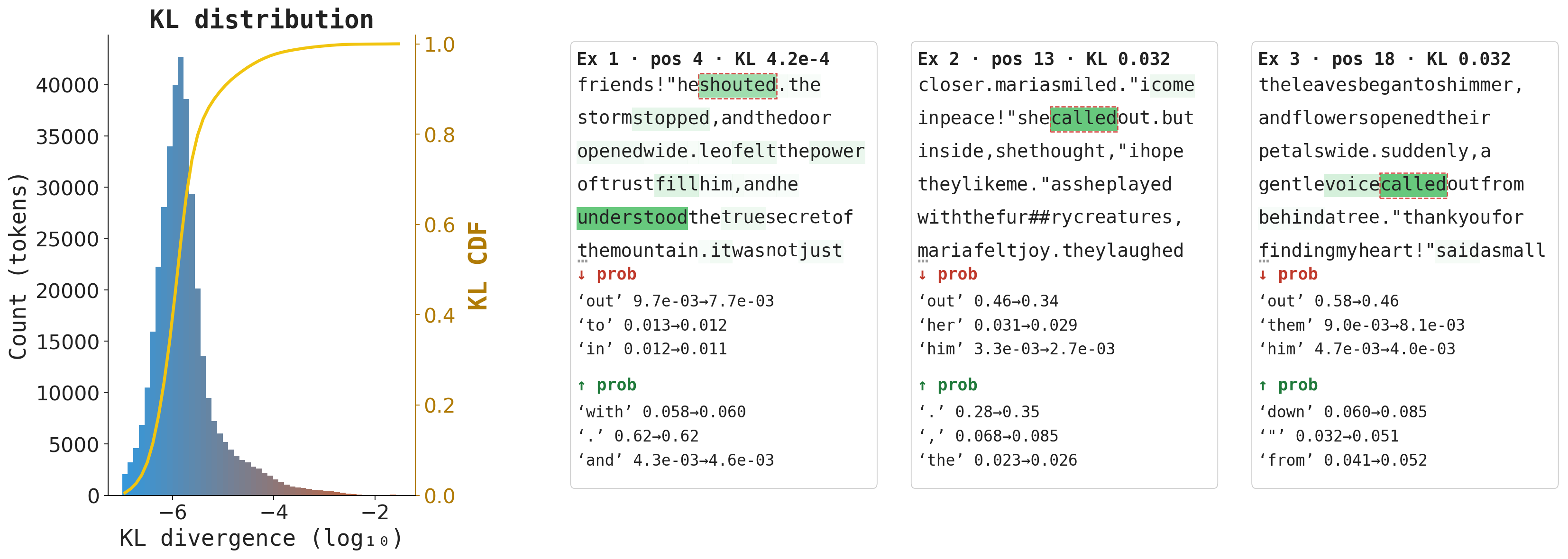}
    \caption{An interpretable weight from \citet{drori2026}. The same display as Figure~\ref{fig:example_gao_a}; the affected tokens cluster around speech-verb contexts (``called'',``shouted'', ``voice'', etc.).}
    \label{fig:example_drori}
\end{figure*}

\section{Methods}

Our pipeline takes a single nonzero weight as input and returns a Python function that describes when the weight is active, together with quantitative scores measuring how well it explains the weight's effect on the model. The pipeline has three stages, summarized in Figure~\ref{fig:procedure}: (i) compute the weight importance location by ablating it and measuring the per-position KL divergence between the original and ablated models; (ii) prompt an LLM to generate candidate predicates from the most-affected token contexts; and (iii) score each candidate by running the original model with the weight gated on or off according to the predicate, and keep the best one. We describe each stage below, give the full algorithm in Algorithm~\ref{alg:per-weight}, and end with the experimental setup.

\subsection{Computing single-weight importance location}
\label{sec:profiling}
To ask whether a single scalar weight $w$ in an MLP layer is interpretable, we first need to know whether and where it matters. We do this by ablation: we run the model twice on the same corpus, once normally and once with $w$ set to zero, and compare the two next-token predictions at every position. The per-position disagreement is the KL divergence between the original and ablated model, which we denote $\mathrm{KL}_p$. The positions with the largest $\mathrm{KL}_p$ are the ones the weight influences most, and the cumulative distribution over $\mathrm{KL}_p$ tells us how concentrated the weight's effect is. We summarize a weight's overall importance by its cross-entropy impact $\Delta\mathrm{CE}$, the difference between the corpus cross-entropy of the ablated model and that of the original. Figures~\ref{fig:example_gao_a},~\ref{fig:example_drori}, ~\ref{fig:example_gao_b} and ~\ref{fig:attn_comment}, show example weights processed this way, with their KL distributions and the token they most affect.

\subsection{Generating candidate predicates}
\label{sec:candidates}

Given the ranked positions from §\ref{sec:profiling}, we want a short, human-readable description of what those positions have in common. We obtain one automatically by selecting the positions accounting for the top 90\% of cumulative KL, using a small window of surrounding tokens around each, and passing the resulting set of contexts to an LLM. The LLM is prompted to return a Python function of the form \texttt{f(tokens, pos) -> bool} together with a one-line natural-language description of what it detects.

In practice the LLM produces predicates ranging from literal token-set membership (\texttt{tokens[pos] in \{"200", "201", "202", ...\}}) to multi-token context tests (\texttt{tokens[pos-1] == '"' and tokens[pos].isalpha()}). The prompt asks for coverage over the affected contexts rather than elegance, and we do not filter or edit the returned code, where we sometimes find minor hallucinations. The context window around each top position is $\pm 8$ tokens. We sample $N$ candidate predicates and choose the best candidate based on the scores detailed in §\ref{sec:scoring}. We sweep $N$ in Appendix~\ref{app:N_sweep} and find our headline numbers saturate around $N = 100$, and we select it as default candidate budget.

\subsection{Scoring a candidate predicate}
\label{sec:scoring}
Given a candidate predicate $f$ for weight $w$, we want to know whether $f$ actually describes what the weight does. The test is a conditional-zero ablation. We globally ablate $w$ throughout the corpus, then restore its per-position contribution only at the positions indicated by $f$, and measure the resulting cross-entropy. Concretely, write $\mathrm{CE}_0$ for the cross-entropy of the unmodified model, $\mathrm{CE}_\varnothing$ for the model with $w$ ablated everywhere, and $\Delta\mathrm{CE} = \mathrm{CE}_\varnothing - \mathrm{CE}_0$ for the ablation effect. Let $\mathrm{CE}_f$ be the cross-entropy of the conditional-zero model that ablates $w$ everywhere except where $f$ holds, and $\mathrm{CE}_{\neg f}$ the symmetric model that ablates $w$ everywhere except where $f$ does \emph{not} hold. We score the predicate by its \emph{recovery} and \emph{inverse}:

\begin{equation}
\mathrm{recovery}(f) \;=\; 1 \;-\; \frac{\mathrm{CE}_f - \mathrm{CE}_0}{\Delta\mathrm{CE}}
\label{eq:rec-inv}
\end{equation}

\begin{equation}
\mathrm{inverse}(f) \;=\; 1 \;-\; \frac{\mathrm{CE}_{\neg f} - \mathrm{CE}_0}{\Delta\mathrm{CE}}.
\label{eq:inv}
\end{equation}

A recovery of $1$ means restoring the weight at the predicted positions fully reproduces the original model, a recovery of $0$ means restoring it there has no effect at all, and a recovery $>1$ can occur on weights with very small $\Delta\mathrm{CE}$ where the conditional-zero model happens to perform marginally better than the unmodified one --- we treat such cases as numerically unreliable and flag them with an asterisk in figures.

Recovery and inverse alone are not enough. A predicate that holds at almost every position trivially recovers most of the ablation effect; in the extreme, the predicate $f \equiv \mathrm{True}$ achieves $\mathrm{recovery}=1$ and $\mathrm{inverse}=0$ while saying nothing about the weight. We rule these out with a \emph{coverage gate}. Writing $p(f)$ for the \emph{coverage} of $f$---the fraction of corpus positions where it fires---we keep a candidate only if it fires on at least one position but no more than half of them, i.e.\ $0 < p(f) \le c$ with coverage cap $c = 0.5$. Among the surviving candidates we summarize recovery and inverse by the interpretability \emph{score},
\begin{equation}
\mathrm{score}(f) \;=\; \min\bigl(\mathrm{recovery}(f),\; 1 - \mathrm{inverse}(f)\bigr),
\label{eq:score}
\end{equation}
and select the highest-scoring candidate $f^*$ among the $N$ returned by the LLM as the weight's interpretation. A weight is called \emph{interpretable at threshold $T$} if its best predicate clears the score threshold and passes the coverage gate,
\[
  \mathrm{score}(f^*) \;\ge\; T \qquad\text{and}\qquad p(f^*) \le c
\]
(equivalently $\mathrm{recovery}(f^*) \ge T$ and $\mathrm{inverse}(f^*) \le 1-T$). We use $T = 0.75$ and $c = 0.5$ throughout the paper and report sensitivity to $T$ in Appendix~\ref{app:llm_comparison}. Algorithm~\ref{alg:per-weight} gives the full pipeline.

\subsection{Models and corpora} \label{sec:setup} We evaluate three training regimes. The weight-sparse code transformer of \citet{gao2025} is trained at $\sim$99\% weight sparsity on Python code. The SimpleStories sparse-and-dense pair of \citet{drori2026} consists of a sparse model trained at $\sim$98\% sparsity on the SimpleStories corpus and a dense counterpart with the same architecture but no sparsity constraint. Pythia-70m \citep{biderman2023} serves as a dense pretrained control, trained on The Pile \citep{Gao2020ThePA}. For each model we evaluate weights on a corpus that approximates its training distribution; see Appendix~\ref{app:method_details} for details. For each model we sample nonzero weights uniformly across MLP layers, and report per-figure sample sizes in the corresponding captions. Unless stated otherwise, the auto-interp LLM $\mathcal{A}$ is Gemini 3 Flash, with the candidate budget set as in §\ref{sec:candidates}. We replicate headline results with Claude Sonnet 4.5, GPT-5, and GPT-4o in Appendix~\ref{app:llm_comparison}. \subsection{Held-out evaluation} \label{sec:heldout} The predicate generation and best-of-n choice both use the same dataset of text, so it might just describe surface patterns of that text rather than the weight itself. To rule this out, we always pick a predicate on one piece of text and \emph{test} it on different text the LLM never saw. Concretely, we take one corpus and cut it into 12 non-overlapping slices from the same distribution. On the first slice we generate candidate predicates and keep the best one, $f^*$ (§\ref{sec:scoring}). We then re-score that same $f^*$ on the remaining held-out slices, which the LLM never saw. This is the whole test: a predicate is only credited if it keeps working on fresh text, not just on the text that produced it, this is what we call $\mathrm{Interp}_B$. Because a single held-out slice gives just one score per weight (and that score depends on which text we happened to draw) we further use the other remaining $K{=}10$ separate held-out slices. Each weight then has ten held-out scores instead of one, and we summarise them by their average and by how much they move from slice to slice (the standard error), i.e.\ how noisy the score is. This gives us two measures that we rely on throughout (§\ref{sec:results-variance}). The \emph{robust} rate counts a weight as interpretable only when we are confident its average held-out score clears the bar, not just on a lucky slice. The \emph{reliability} says how much of the spread in scores across a model's weights reflects a real property of the weights, rather than noise from which text we happened to test on. Both are stricter than a single check, and both test the one thing we care about: whether a predicate works off the corpus that produced it.

%% file: sections/2_results.tex
\section{Results}
\label{sec:results}

\begin{figure*}
\centering \includegraphics[width=\textwidth]{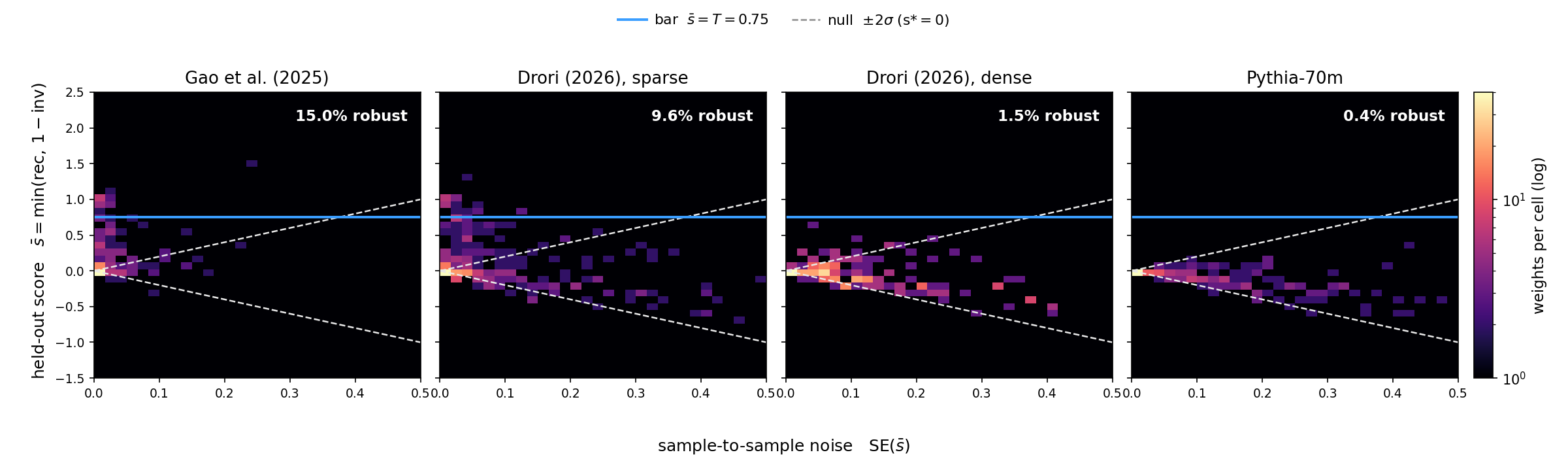} \caption{Robust interpretability on held-out data. Each weight is re-scored on $K{=}10$ disjoint held-out slices: the vertical axis is its mean score $\bar s = \min(\mathrm{rec},\,1-\mathrm{inv})$ and the horizontal axis is how much that score varies between slices (its standard error). Color is the log density of weights, showing the \emph{robustly interpretable} weights above the interpretability bar (blue, $\bar s \geq 0.75$) \emph{and} outside the ``not interpretable'' $\pm2\sigma$ null cone (dashed). The percentage is the robust rate per model.} \label{fig:robust_scatter} \end{figure*}

\begin{table}[h] \centering \tiny \caption{Pooled per-weight interpretability rates. Each entry is the percentage of a model's sampled nonzero weights that pass the test ($\pm$1~SEM%
). \textbf{Interp$_A$}: the weight's predicate is scored on the same text it was written from (in-sample). \textbf{Interp$_B$}: the same predicate, scored instead on separate held-out text. \textbf{Robust}: the predicate is re-scored on $K{=}10$ held-out slices, and the weight counts only if its average held-out score is both clearly interpretable and reliably above a ``not-interpretable'' baseline.}\label{tab:pooled} \begin{tabular}{l cc c} \toprule Model & $\mathrm{Interp}_A$ & $\mathrm{Interp}_B$ & $\mathrm{Robust}$\\ \midrule Gao et al.\ (2025)     & 37.3~$\pm$~3.9\% & 35.3~$\pm$~3.9\% & \textbf{15.0~$\pm$~1.7\%} \\ Drori (2026), sparse   & 21.4~$\pm$~1.5\% & 17.2~$\pm$~1.4\% & \textbf{9.6~$\pm$~1.1\%}  \\ Drori (2026), dense    & 13.2~$\pm$~1.4\% & 9.0~$\pm$~1.2\%  & \textbf{1.5~$\pm$~0.6\%}  \\ Pythia-70m             & 12.0~$\pm$~0.9\% & 5.0~$\pm$~0.6\%  & \textbf{0.4~$\pm$~0.3\%}  \\ \bottomrule \end{tabular} \end{table}

\subsection{Sparse models are significantly more interpretable than comparable dense models} \label{sec:results-headline} On a per-weight basis, the sparse models are far more interpretable than dense models of comparable size. Throughout we summarise a model by its \emph{robust rate}: the fraction of weights whose interpretation holds up on held-out text, credited only when a weight's average score across the $K{=}10$ held-out slices reliably clears the interpretability bar (§\ref{sec:heldout}). At the canonical settings ($T=0.75$, coverage cap $c=0.5$), the two sparse models sit far above the dense controls --- roughly $15\%$ and $10\%$ of their non-zero weights are robustly interpretable, against about $1\%$ for the dense model and essentially none for the pretrained dense control (Table~\ref{tab:pooled}). This ordering is not an artifact of one threshold or one judge: it holds within every impact bin below and survives every robustness check we run (Appendices~\ref{app:N_sweep}--\ref{app:llm_comparison}).

Where do the interpretable weights sit? Binning each model's weights by the size of their causal effect $|\Delta\mathrm{CE}|$ (Figure~\ref{fig:ce_distribution}) shows two things. The sparse models place a real population of weights in the high-impact range ($|\Delta\mathrm{CE}| \geq 10^{-4}$), and that is exactly where interpretability is densest --- roughly a quarter of their high-impact weights are robustly interpretable. The dense models barely populate that range at all, and what little they place there generalises poorly; the pretrained control reaches no high-impact weights in our sample. The right-most column of Figure~\ref{fig:ce_distribution} ($=0$) counts parameters that are exactly zero by construction, and is excluded from all rates.

\paragraph{What the $|\Delta\mathrm{CE}|$ range means.} Per-weight ablation effects span four orders of magnitude in all four models, from above $10^{-3}$ at the high end to below $10^{-6}$ at the low end. The lowest bin ($<10^{-6}$) is at the noise floor of the cross-entropy measurement and contributes essentially nothing to the model's loss; weights there can score high on a predicate by accident, since any predicate's effect on a near-zero $\Delta\mathrm{CE}$ is dominated by numerical noise rather than the weight's actual function. The coverage gate ($p \leq c = 0.5$ in Algorithm~\ref{alg:per-weight}) rules out the most flagrant version of this failure mode --- predicates that fire almost everywhere and ``recover'' the ablation by trivially restoring the weight at most positions --- but the noise floor remains the reason the lowest CE bin is not where the interpretable population lives. The higher-impact bins are where sparse and dense models part ways: a sparse model places significant probability mass there and most of it is interpretable; a dense model places little mass there and what it does place generalises worse off-corpus.

\begin{figure}[t]
    \centering
    \includegraphics[width=0.52\textwidth]{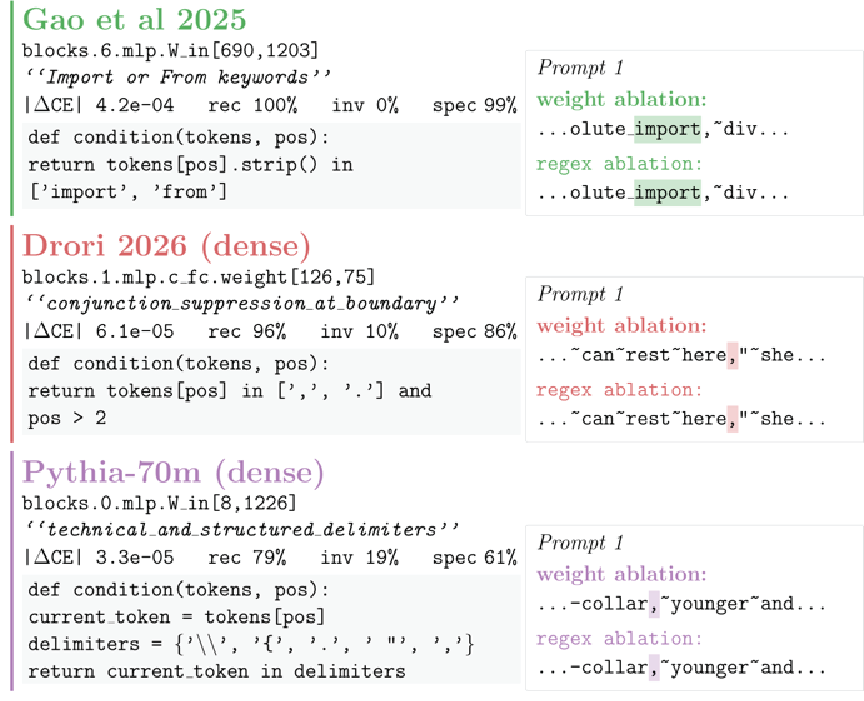}
    \caption{Example weights and their LLM-generated predicates. Each row gives a weight's identifier and predicate code, with a high-KL prompt comparing where the weight fires (weight ablation) against where the predicate evaluates true (regex ablation). Rows show, top to bottom, a Gao-sparse, a Drori-dense, and a Pythia-70m weight.}
    \label{fig:regex}
\end{figure}

\subsection{Sparse-model scores are properties of the weight, not the corpus} \label{sec:results-variance} The rates above rest on held-out text, but we can ask a sharper question of the same data: not \emph{how many} weights pass, but \emph{how much} each weight's score depends on which held-out sample it happened to be measured on. If a score reflects the weight, redrawing the sample should barely move it; if it reflects the corpus, it should change.

Re-scoring every weight on $K{=}10$ disjoint held-out splits separates the two. The spread of scores across a model's weights mixes genuine differences between weights with sample-to-sample wobble in each weight's own score; subtracting the wobble leaves the part that survives resampling, whose share of the total we call the \emph{reliability} (Figure~\ref{fig:reliability}). It falls steeply from the sparse models to the dense ones: on the sparse models almost all of the spread is real (around $85$--$90\%$), so a weight's score is essentially a stable property of that weight; on the pretrained dense control barely half survives ($47\%$), meaning most of what looks interpretable there is decided by which split happened to be drawn.

This instability does more than add noise --- it inflates the dense models' apparent interpretability, and removing it makes them collapse. Our headline robust rate keeps a weight only if its held-out score clears the bar \emph{and} is confidently above a ``not-interpretable'' baseline, not merely on a lucky split (Figure~\ref{fig:robust_scatter}). Under this test the sparse models retain a sizeable interpretable population while the dense controls fall to near zero (Table~\ref{tab:pooled}): a single held-out check had credited the dense models with a few percent, but almost none of it survives resampling. 

Every held-out measure tells the same story (Table~\ref{tab:pooled}): the held-out pass rate, the reliability, and the robust rate all give the same ordering, and correcting for sampling noise \emph{widens} the sparse--dense gap rather than narrowing it. The sparse models carry interpretability that belongs to their weights; the dense models' belongs largely to the corpus it was measured on.

\begin{figure*}
    \centering
    \includegraphics[width=0.9\textwidth]{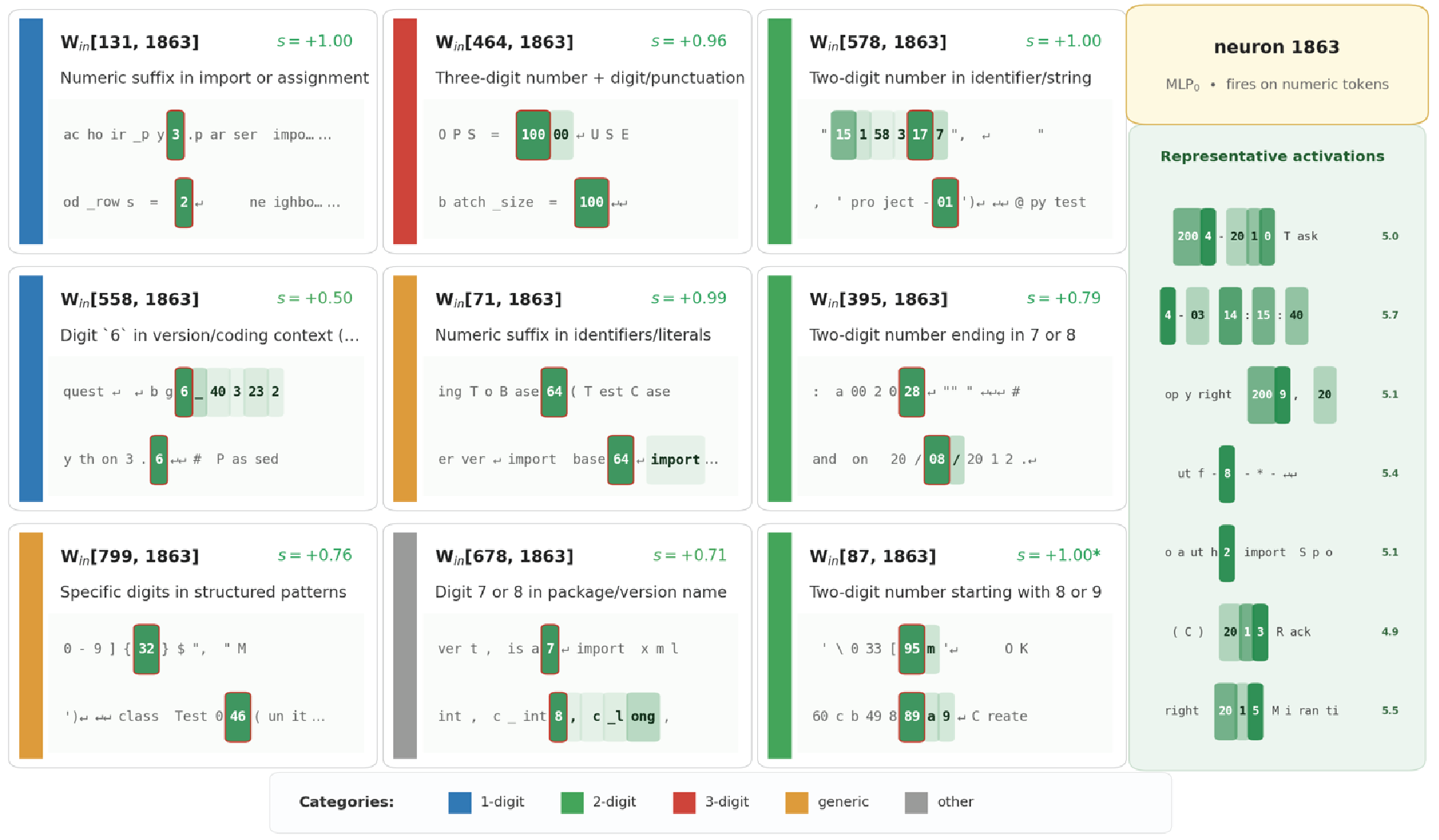}
    \caption{A neuron in the \citet{gao2025} model that fires on numeric tokens. Each of the neuron's nine nonzero input weights is labelled with the LLM-generated interpretability score. {Note: * means that $|\Delta\mathrm{CE}| < 0.00001$}.}
    \label{fig:case-digit}
\end{figure*}

\subsection{What the predicates look like.} The LLM-generated predicates cluster into a few structural patterns across all four models. The most common form is a single-token lookup: ``is the focus token a digit,'' ``is it one of $\{\texttt{th}, \texttt{ode}, \texttt{b}\}$,'' ``is it a mid-word lowercase fragment.'' A smaller fraction conditions on the immediate neighbour: ``previous word ends in \texttt{-ed},'' ``next token is an inflectional suffix.''\footnote{These phrases are the auto-interp LLM's own one-line descriptions; each is paired with an executable Python predicate (e.g.\ \texttt{tokens[pos] in \{"ed","ing","s","es","ly"\}} for the inflectional-suffix example.} Even the longer predicates rarely reach beyond two or three adjacent tokens, and we do not observe predicates that maintain state across a prompt (e.g.\ ``inside an open string literal''). The pattern is consistent with weights that act locally in token-context, even when the higher-level circuits they participate in are not. Figure~\ref{fig:regex} shows three representative weights (one Gao sparse, one Drori dense, one Pythia) with the predicate code and a few high-KL prompts where the weight fires; predicate length and further class statistics across all four models are in Appendix~\ref{app:rec_inv_distributions}.

\subsection{Threshold sensitivity and pipeline ablations}
\label{sec:results-robustness}

The interpretable rates reported above depend on a threshold $T$ that is inherent to the metric and on two pipeline hyperparameters --- the choice of LLM and the number of candidate predicates $N$ generated per weight. We treat these separately: $T$ is a core property of the interpretability measure, whereas the LLM and $N$ are knobs of our evaluation pipeline that we ablate to confirm the cross-model ordering does not depend on them.

\textbf{Threshold sensitivity.} The interpretable rate depends on the predicate-pass threshold $T$ used to convert per-weight scores into a binary interpretable/not-interpretable judgment. As $T$ varies, absolute rates shift monotonically, but the cross-model ordering (Gao $>$ Drori sparse $>$ Drori dense $>$ Pythia) is preserved across the full sweep we tested (Appendix~\ref{app:llm_comparison}). 

\textbf{LLM choice and candidate budget.} The choice of LLM and the number of LLM calls per weight $N$ are hyperparameters of our pipeline rather than properties of the underlying weights. We re-run the full evaluation under four different LLMs (Gemini 3 Flash, Claude Sonnet 4.5, GPT-5, GPT-4o): absolute rates at $T=0.75$ differ by up to $\sim$30 percentage points, with stronger LLMs higher, but the rate-vs-$T$ shape and the cross-model rank order are preserved under each LLM (Appendix~\ref{app:llm_comparison}). Rates also rise with $N$ and saturate by $N \approx 100$ on the sparse models, while the dense baselines stay low across the entire sweep $N \in \{1, 3, 5, 10, 20, 100, 150\}$ (Appendix~\ref{app:N_sweep}). The cross-model ordering Gao $>$ Drori sparse $>$ Drori dense $>$ Pythia therefore holds under each LLM and under each candidate budget.

%% file: sections/2b_casestudies.tex
\section{Case Studies}
\label{sec:case-studies}
To ground the rates of §\ref{sec:results} in concrete examples, we work through one MLP neuron of the \citet{gao2025} sparse code transformer. The neuron fires predominantly on numeric tokens at the activation level; applied to its nine nonzero input weights, our pipeline assigns a separately-interpretable predicate to each, and the predicates cluster into multiple digit-count categories --- single-digit, two-digit, three-digit, generic and other gates that together compose the activation-level ``detects digits'' label (Figure~\ref{fig:case-digit}). The full decomposition is in Appendix~\ref{app:digit_neuron}, and two further worked examples are in Appendices~\ref{app:case_quotes} (string-closing context dependence) and~\ref{app:case_drori_speech} (a Drori-sparse speech-verb neuron decomposed into five functional sub-roles).

%% file: sections/3_conclusion.tex
\section{Conclusion}
\label{sec:conclusion}
We find that a substantial fraction of nonzero weights in weight-sparse transformers admit individually interpretable explanations of when they are active, whereas the corresponding fraction in dense models is nearly zero. The cross-model ordering is preserved across thresholds, candidate budgets, and choice of auto-interp LLM. The interpretable rate also seems to depend on a weight's causal impact. Weights with larger ablation effects are more likely to admit explanations our pipeline can recover, while weights with very small ablation effects are interpreted less reliably, suggesting that low-impact weights either contribute too little signal for the auto-interp LLM to characterize or play roles that the current predicate language cannot express. Three case studies show what the aggregate rates correspond to in practice. We find per-digit-count gates inside a digit-detector neuron, look-back-dependent functions inside the string-closing circuit, and a speech-verb neuron that decomposes at the weight level into five parallel functional roles (speech-verb activators, mental-verb suppressors, a modal suppressor, punctuation gates, and specific-token gates). Combined with the trivial interpretability of the zero weights, the great majority of parameters in these sparse models admit per-weight explanations, while the same is not true of dense models.

\paragraph{Limitations.} Our measurements tell us \emph{when} a weight is active, not \emph{what} it computes once active. A function that correctly predicts the inputs on which a weight matters is informative, but it is not a complete account of the weight's role in the model. The numbers we report should be read as evidence that weights in sparse transformers are individually identifiable functional units, not as a complete characterization of what each one does. Another limitation is that our numbers are a lower bound on what the method could in principle find. Our function language is Python predicates over token context, which cannot express semantic features (e.g., whether a span refers to a person) and cannot easily express features that span multiple tokens or require state across the prompt. A weight whose role depends on such structure would be filed as not interpretable here, even if it has a clean account of when it matters that our language simply cannot write down. %
Finally, our held-out corpus is drawn from the same pretraining distribution as the evaluation corpus; whether these descriptions survive distribution shift is a separate question out of scope in this paper.

\paragraph{Future work.} Three extensions follow naturally. First, the function language could be extended beyond token-local tests (for instance to predicates that reference parser state, BPE-segmentation features, or the outputs of learned probes) which should recover weights whose domain of action the current language cannot describe. Second, the fraction of robustly interpretable weights is a quantity one can train against, prune by, or use to compare architectures, in the role that token-level perplexity plays for capability. Third, applying the same method to larger sparse-trained transformers (ideally ones with non-trivial multi-step behavior) will reveal whether per-weight interpretability is a property of these particular small models or of sparsity training itself. 

%% file: sections/4_appendix.tex
\newpage 
\section{Appendix}

\subsection{Corpus Selection Additional Details}
\label{app:method_details}

We use CodeParrot \citep{codeparrot_dataset} as a proxy for the original Python pretraining data of the \citet{gao2025} model, which is not publicly released, and stream The Pile \citep{Gao2020ThePA} for Pythia-70m, which matches the pretraining distribution but not the exact training sequences. For the Drori (2026) sparse and dense models we use SimpleStories \citep{drori2026}, which is the actual training corpus.

To verify that our chosen corpora are reasonable approximations of each model's training distribution, we compare the cross-entropy loss of each model on its evaluation corpus against the loss on deliberately mismatched corpora. Table~\ref{tab:corpus_match} reports the comparison. The \citet{gao2025} model achieves a CE of 2.34 on CodeParrot, but its loss rises sharply on out-of-distribution text. SimpleStories English yields a CE of 3.70, and French Wikipedia yields 4.50, both substantially higher than the in-distribution CE. Likewise, the Drori sparse model achieves a CE of 2.98 on SimpleStories, well below what either model achieves outside its training domain. We take the size of the in-distribution-versus-out-of-distribution gap as evidence that CodeParrot and SimpleStories are close enough to each model's training distribution to serve as evaluation corpora for our pipeline.

\begin{table}[h]
\centering
\small
\caption{Cross-entropy loss of each model on in-distribution and out-of-distribution corpora. The large gap between in- and out-of-distribution CE indicates that our chosen evaluation corpora are reasonable approximations of each model's training distribution.}
\label{tab:corpus_match}
\begin{tabular}{l l c c}
\toprule
Model & Corpus & CE Loss & In-dist? \\
\midrule
Gao et al.\ (2025)   & CodeParrot (Python)        & 2.34 & Yes \\
Drori (2026), sparse & SimpleStories              & 2.98 & Yes \\
Gao et al.\ (2025)   & SimpleStories (English)    & 3.70 & No  \\
Gao et al.\ (2025)   & French Wikipedia           & 4.50 & No  \\
\bottomrule
\end{tabular}
\end{table}

The per-weight ablation effect $|\Delta\mathrm{CE}|$ spans multiple orders of magnitude across all models, from $\sim\!10^{-3}$ down to below $10^{-6}$ (Figure~\ref{fig:ce_distribution}).

\begin{figure}[h]
    \centering
    \includegraphics[width=\textwidth]{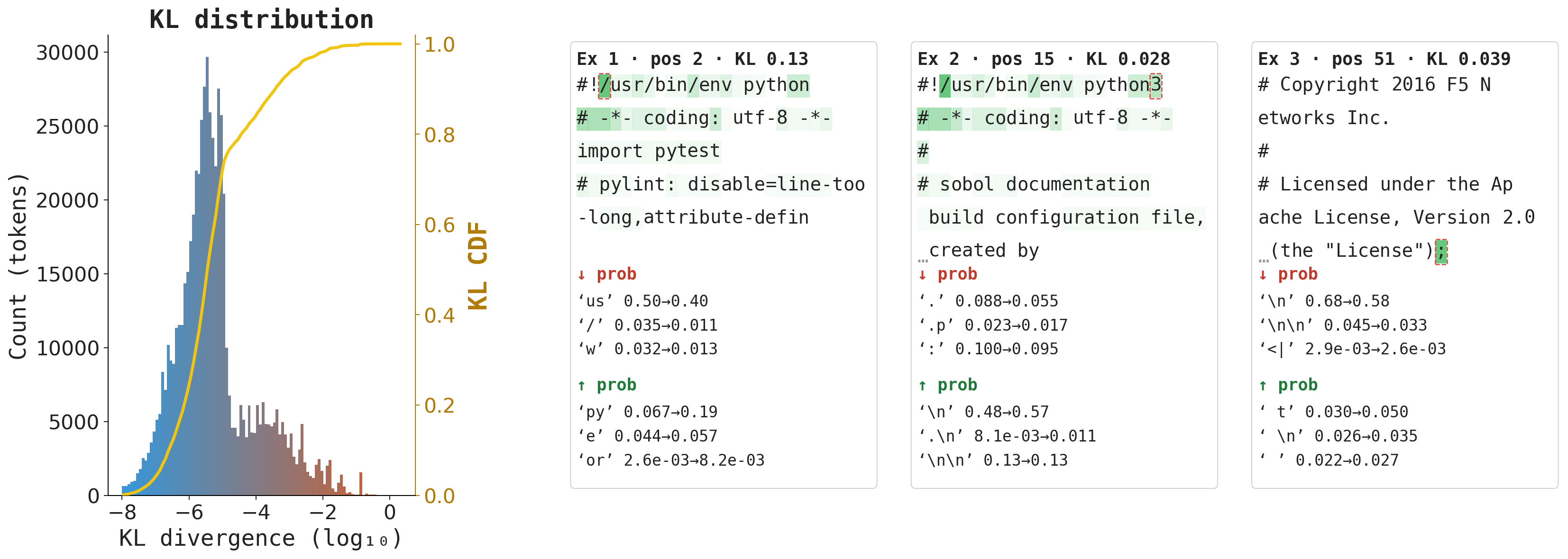}
    \caption{Another representative MLP weight from \citet{gao2025} sparse code transformer.}
    \label{fig:example_gao_b}
\end{figure}

\begin{figure}[t] \centering \includegraphics[width=\textwidth]{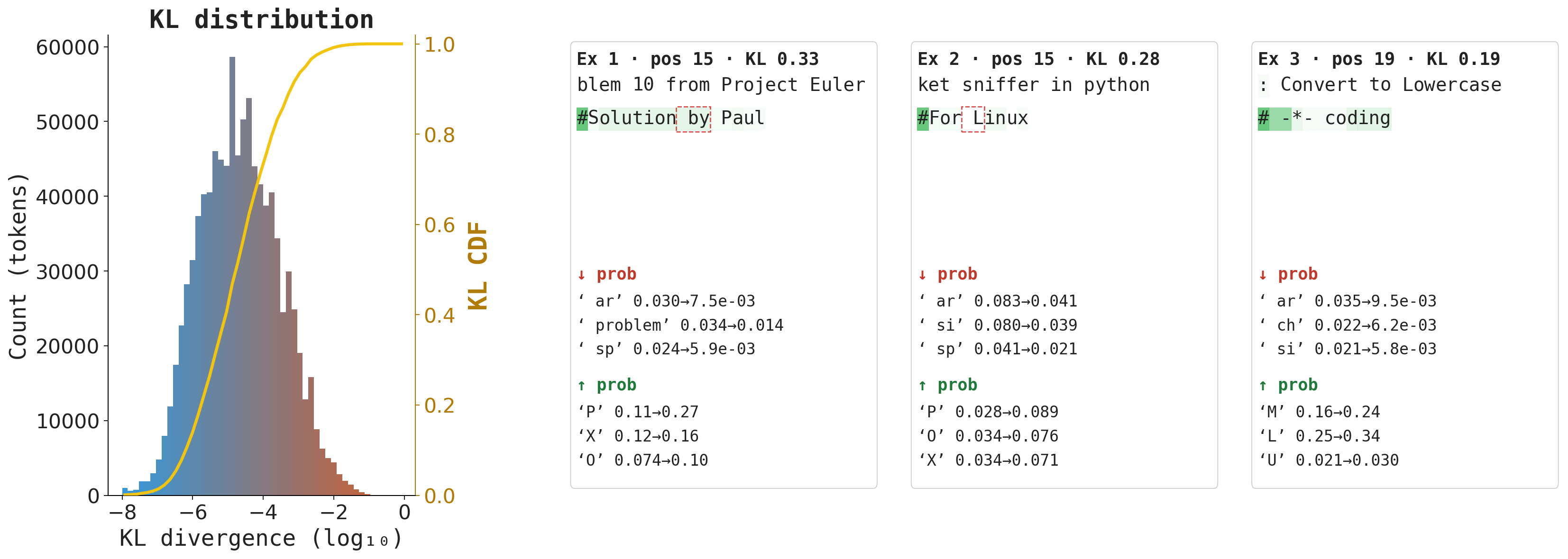} \caption{Representative interpretable attention weight from the \citet{gao2025} sparse code transformer.} \label{fig:attn_comment} \end{figure}

\subsection{Distributions of recovery and inverse}
\label{app:rec_inv_distributions}

\begin{algorithm}
\caption{Per-weight conditional-zero pipeline.}
\label{alg:per-weight}
\textbf{Input:} Model $M$, weight $w$, corpus $X$, LLM $\mathcal{A}$, cap $c$. \\
\textbf{Output:} Best predicate $f^*$ and its scores $(r^*, v^*, p^*, s^*)$.
\begin{algorithmic}[1]
\State \textbf{Compute the weight's effect across the corpus.} Run $M$ and $M_{w=0}$ on $X$.
\State \quad Compute $\Delta\mathrm{CE} \gets \mathrm{CE}_\varnothing - \mathrm{CE}_0$, the corpus-level cross-entropy gap.
\State \quad Compute $\mathrm{KL}_p$ at every position; keep the top $90\%$ of cumulative KL positions.

\smallskip
\State \textbf{Propose candidates.} Pass the selected positions to $\mathcal{A}$ and sample $K$ predicates $f_1, \dots, f_K$.

\smallskip
\State \textbf{Score each candidate.}
\For{$k = 1, \dots, K$}
  \State $p_k \gets \dfrac{1}{|X|}\sum_{x \in X}[f_k(x)]$
         \Comment{\emph{coverage}: fraction of positions where $f_k$ fires}
  \If{$p_k = 0\ \textbf{or}\ p_k = 1\ \textbf{or}\ p_k > c$}
    \State $s_k \gets -\infty$ \Comment{trivially empty / broad: ruled out}
    \State \textbf{continue}
  \EndIf
  \State Build $M_{f_k}$: ablate $w$ everywhere, restore it where $f_k$ fires.
  \State Build $M_{\neg f_k}$: ablate $w$ everywhere, restore it where $f_k$ does \emph{not} fire.
  \State $r_k \gets 1 - \dfrac{\mathrm{CE}_{f_k} - \mathrm{CE}_0}{\Delta\mathrm{CE}}$
         \Comment{\emph{recovery}: effect explained by restoring at $f_k$'s firing positions}
  \State $v_k \gets 1 - \dfrac{\mathrm{CE}_{\neg f_k} - \mathrm{CE}_0}{\Delta\mathrm{CE}}$
         \Comment{\emph{inverse}: effect explained by the predicate's complement}
  \State $s_k \gets \min\!\left(r_k,\; 1 - v_k\right)$
         \Comment{\emph{interpretability score}}
\EndFor

\smallskip
\State \Return $f^* \gets f_{k^*}$ where $k^* = \arg\max_k s_k$, and its scores $(r^*, v^*, p^*, s^*)$.

\smallskip
\Statex \textbf{Interpretability decision (threshold $T$).} A weight is called \emph{interpretable} if
\[
  r^* \geq T \quad\textbf{and}\quad v^* \leq 1-T \quad\textbf{and}\quad p^* \leq c.
\]
Default settings used: $T=0.75$ and $c=0.5$.
\end{algorithmic}
\end{algorithm}

\begin{figure}[t]
    \centering
    \includegraphics[width=\textwidth]{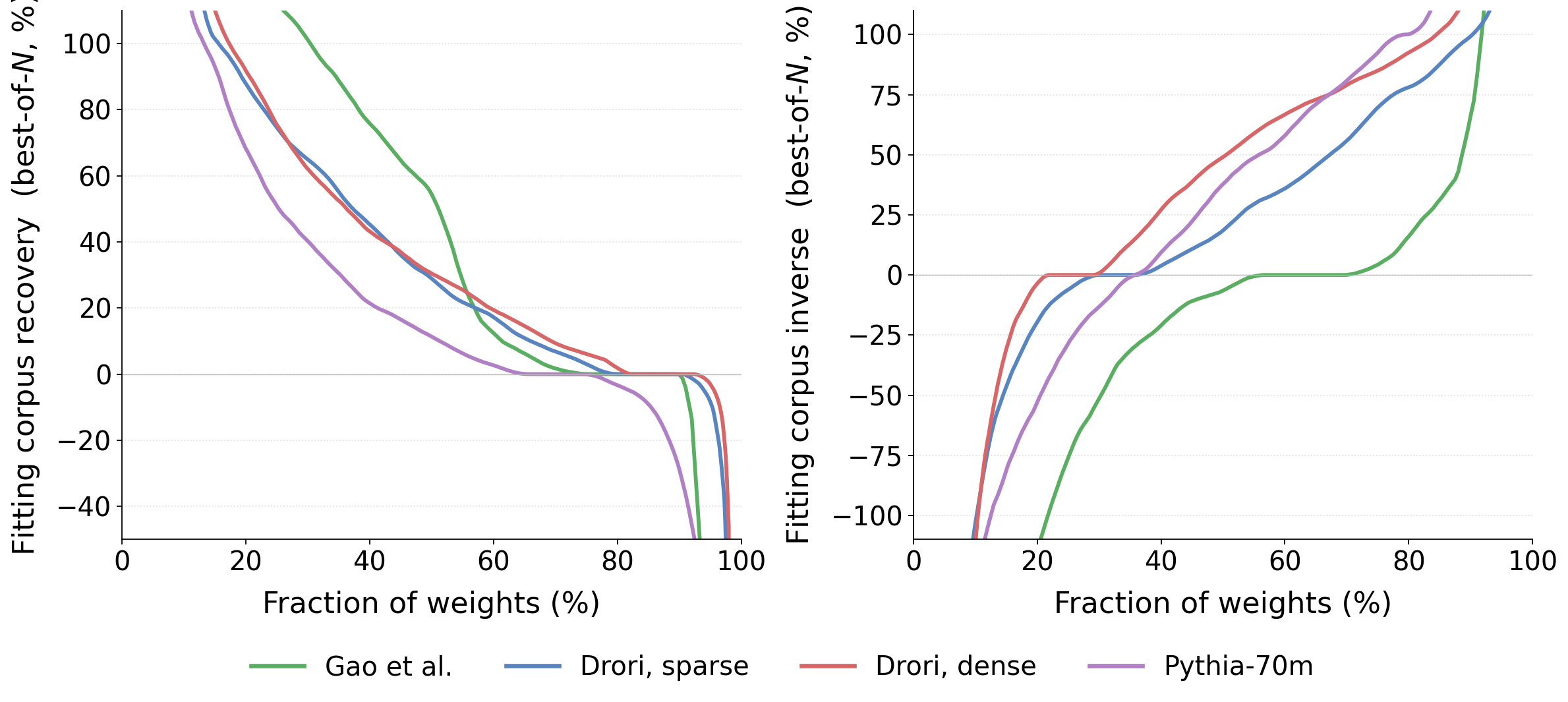}
    \caption{Best-of-$N$ recovery (left) and inverse (right) across all sampled weights, sorted within each model so curves are monotone. Dashed lines mark the canonical thresholds.}
    \label{fig:rec_inv}
\end{figure}

\begin{figure}[t]
    \centering
    \includegraphics[width=\textwidth]{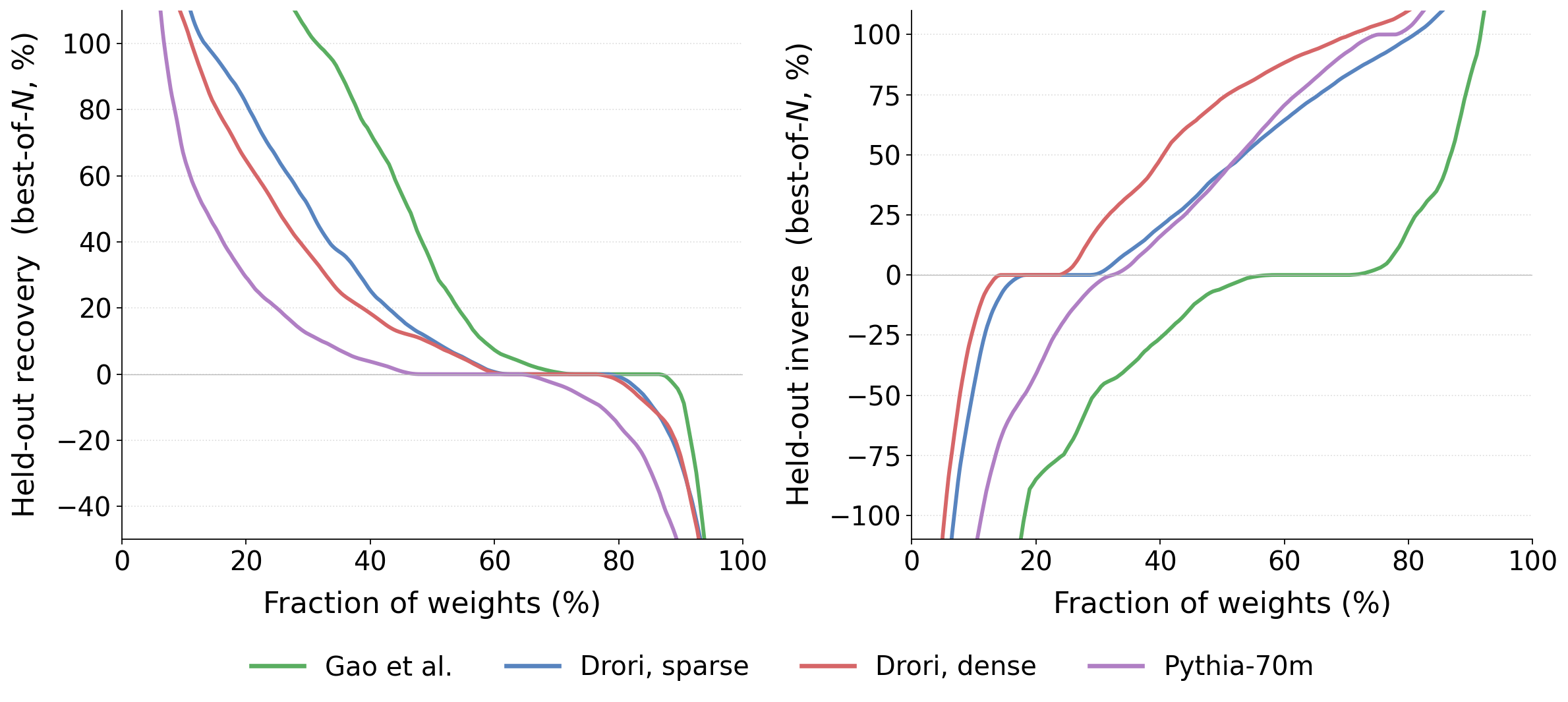}
    \caption{Best-of-$N$ recovery (left) and inverse (right) on the held-out corpus. Same setup as Figure~\ref{fig:rec_inv}, but evaluated on a corpus never seen when generating the candidate predicates.}
    \label{fig:rec_inv_heldout}
\end{figure}

Figures~\ref{fig:rec_inv} and \ref{fig:rec_inv_heldout} show the distributions of best-of-$N$ recovery and inverse scores across all sampled weights, sorted within each model. These curves give a more complete view of the behavior summarized by the headline interpretability rates. The sparse models have a long tail of weights that pass both the recovery and inverse criteria, while the dense baselines reach the canonical thresholds only on a much smaller slice. Negative inverse values for the sparse models reflect predicates whose negation raises loss above the fully-ablated baseline, which is direct evidence that the predicate is identifying a real, localized effect rather than partitioning a uniform contribution.

Comparing Figure~\ref{fig:rec_inv} to the held-out version in Figure~\ref{fig:rec_inv_heldout}, the gap between the sparse and dense models widens. The sparse models retain most of their best-of-$N$ recovery on the held-out corpus, while the dense models lose more of theirs, which suggests that predicates fitted to sparse-model weights generalize better than those fitted to dense-model weights. We read this as further evidence that the predicates for sparse weights are tracking a real property of the weight rather than overfitting to features of the corpus used to generate them.

\begin{figure}[t]
    \centering
    \includegraphics[width=0.85\textwidth]{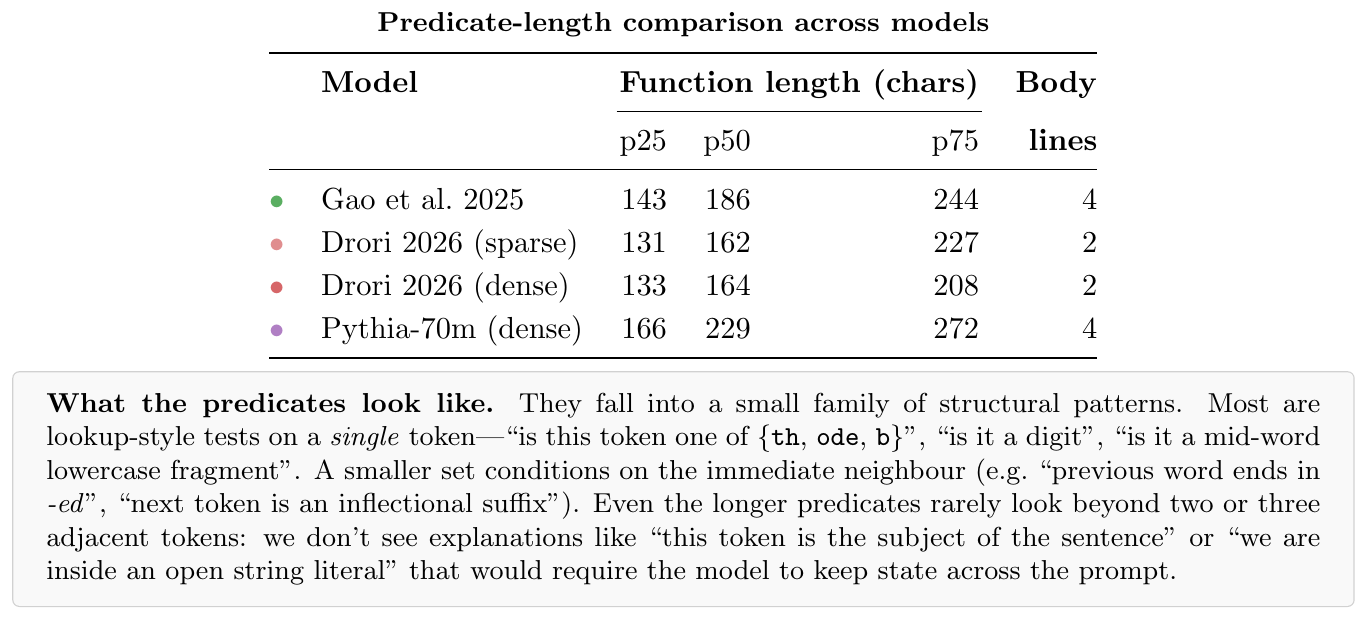}
    \caption{Structural statistics of the LLM-generated predicates per model: distribution of predicate length (lines of code) and predicate class (single-token lookup, neighbor test, multi-token context test).}
    \label{fig:regex_lengths}
\end{figure}

Figure~\ref{fig:regex_lengths} reports structural statistics of the predicates themselves. Predicates remain short across all models (median $\leq 4$ lines, and only $2$ lines on the Drori sparse and dense models), and the predicate language is the same on sparse and dense models. The difference in interpretability rates is therefore not driven by sparse models receiving longer or more elaborate predicates.

\subsection{Marginal shape of $s_B$ vs.\ per-sample consistency}
\label{app:variance_vs_shape}
This appendix decomposes the gap and explains why we treat the marginal-shape effect as the dominant driver. Throughout, $s_A$ denotes a predicate's score on the text it was generated from and $s_B$ its score on held-out text the LLM never saw.

\begin{figure}[h] \centering \includegraphics[width=\textwidth]{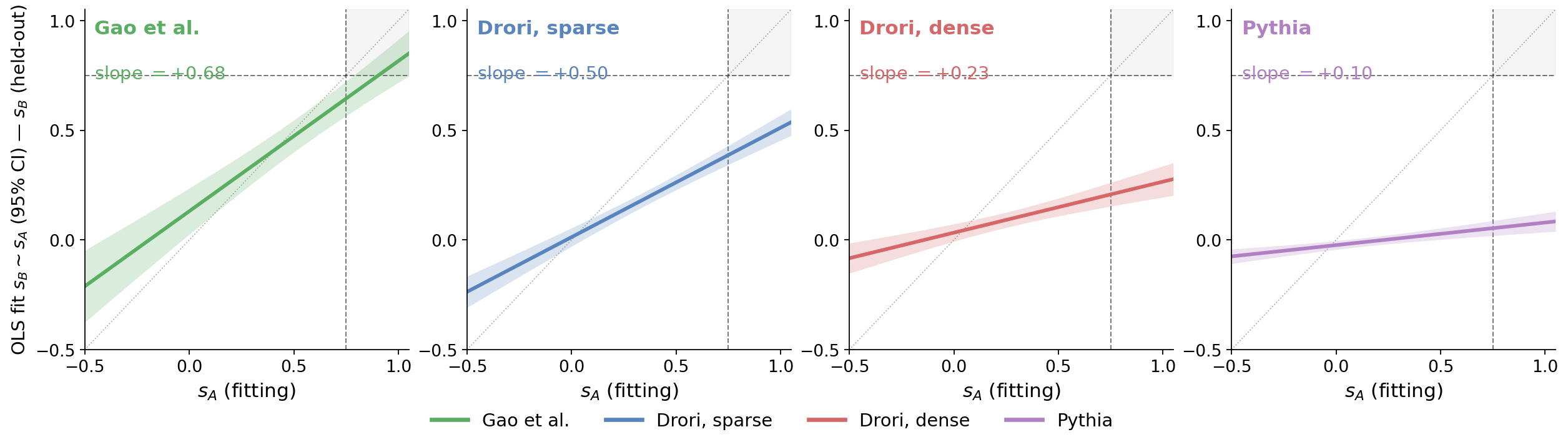} \caption{Held-out interpretability score $s_B = \min(\mathrm{rec}_B,\,1-\mathrm{inv}_B)$ across models. OLS fit $s_B \sim s_A$ per model with $95\%$ CI; slope annotated.} \label{fig:variance} \end{figure}

\paragraph{Per-sample consistency.} Figure~\ref{fig:variance} fits an ordinary least-squares regression $s_B \sim s_A$ per model and reports the slope with a 95\% CI band. A slope of $1$ would indicate that the held-out score perfectly tracks the generating-text score; a slope of $0$ would indicate that $s_A$ has no predictive content for $s_B$ and a predicate's apparent score is accidental. The fitted slopes order in the same direction as the headline rates.

\begin{figure}[h] \centering \includegraphics[width=0.85\textwidth]{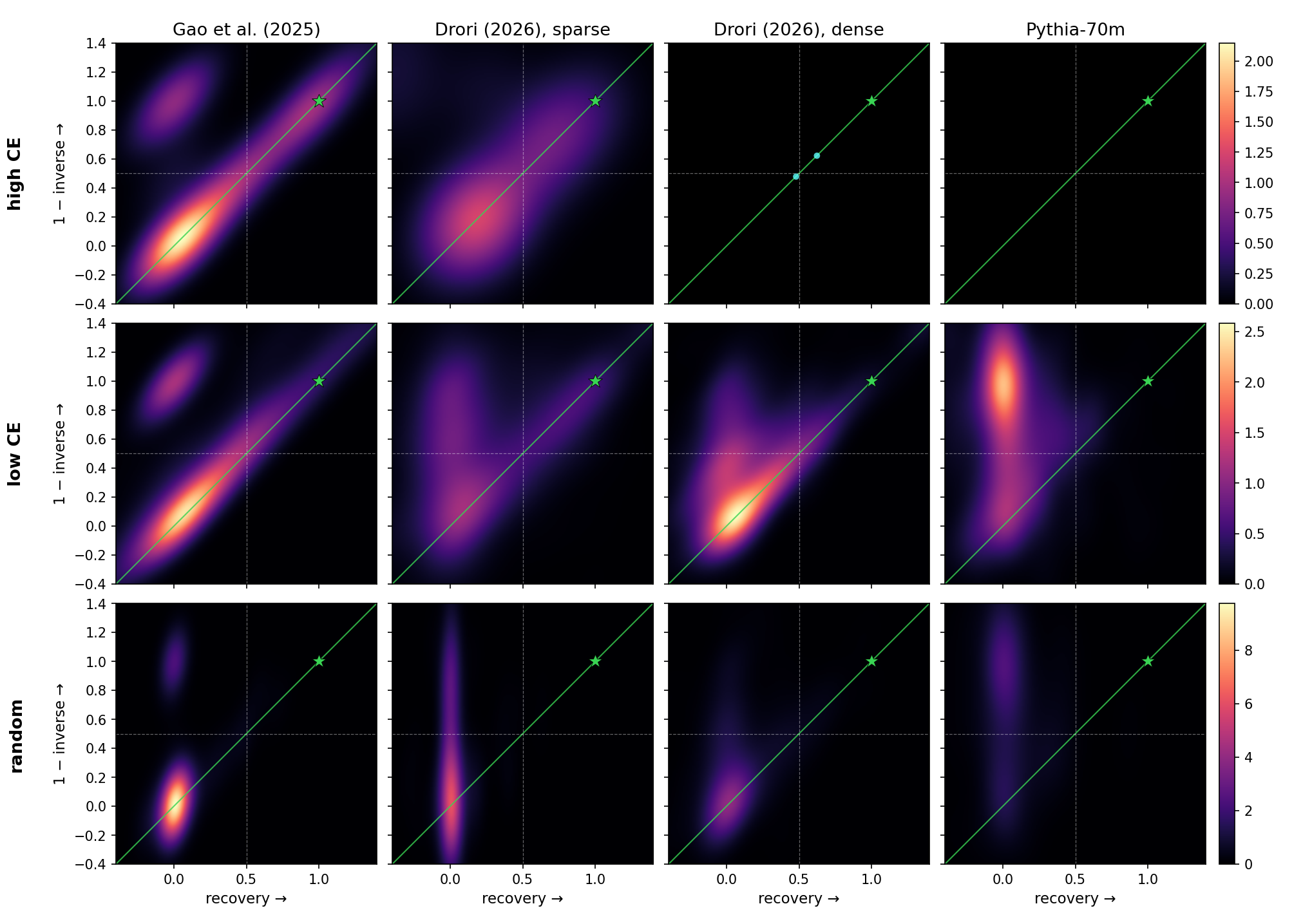} \caption{Reliability of per-weight interpretability scores under held-out resampling. Each weight is re-scored on $K{=}10$ disjoint held-out splits; bars show the fraction of each model's cross-weight score variance that survives redrawing the corpus ($95\%$ bootstrap CIs). Sparse-model scores are mostly stable properties of the weight; on Pythia, most of the apparent spread is sampling noise.} \label{fig:reliability} \end{figure}

\paragraph{Random-chance baseline.} A useful way to see why the marginal effect dominates is to ask what interpretability scores a random predicate would produce. Consider a predicate that fires on a uniformly random subset of corpus positions with coverage $p_{\mathrm{fire}}$, independent of the weight's actual firing pattern. Restoring the weight at those positions recovers approximately $p_{\mathrm{fire}}$ of the ablation effect (rec $\approx p_{\mathrm{fire}}$); restoring at the complement recovers approximately $1 - p_{\mathrm{fire}}$ (inv $\approx 1 - p_{\mathrm{fire}}$); and the score is $s \approx \min(p_{\mathrm{fire}},\,p_{\mathrm{fire}}) = p_{\mathrm{fire}}$. Under the gate ($p_{\mathrm{fire}} \leq 0.5$) the random-chance score is bounded above by $0.5$, well below the interpretability threshold $s \geq 0.75$. A weight whose predicate scores high on the generating text purely by chance therefore sits near $0.5$, and re-scoring on held-out text pulls it toward the random-chance mean rather than near $1$. Dense-model weights contribute more mass near $0.5$ and less mass near $1$, so under held-out re-scoring they regress toward the random-chance mean and lose more than sparse-model weights do. This is why evaluating on held-out text widens the sparse--dense gap rather than narrowing it, and why the widening is mostly attributable to this marginal-shape difference rather than to a per-weight noise difference. The same effect is what the $K{=}10$ resampling in Figure~\ref{fig:reliability} measures directly: subtracting sample-to-sample wobble leaves the dense models with little stable interpretable mass, since most of theirs sat near the random-chance region to begin with.


\subsection{Sensitivity to the candidate budget}
\label{app:N_sweep}

\begin{figure}[h]
    \centering
    \includegraphics[width=\textwidth]{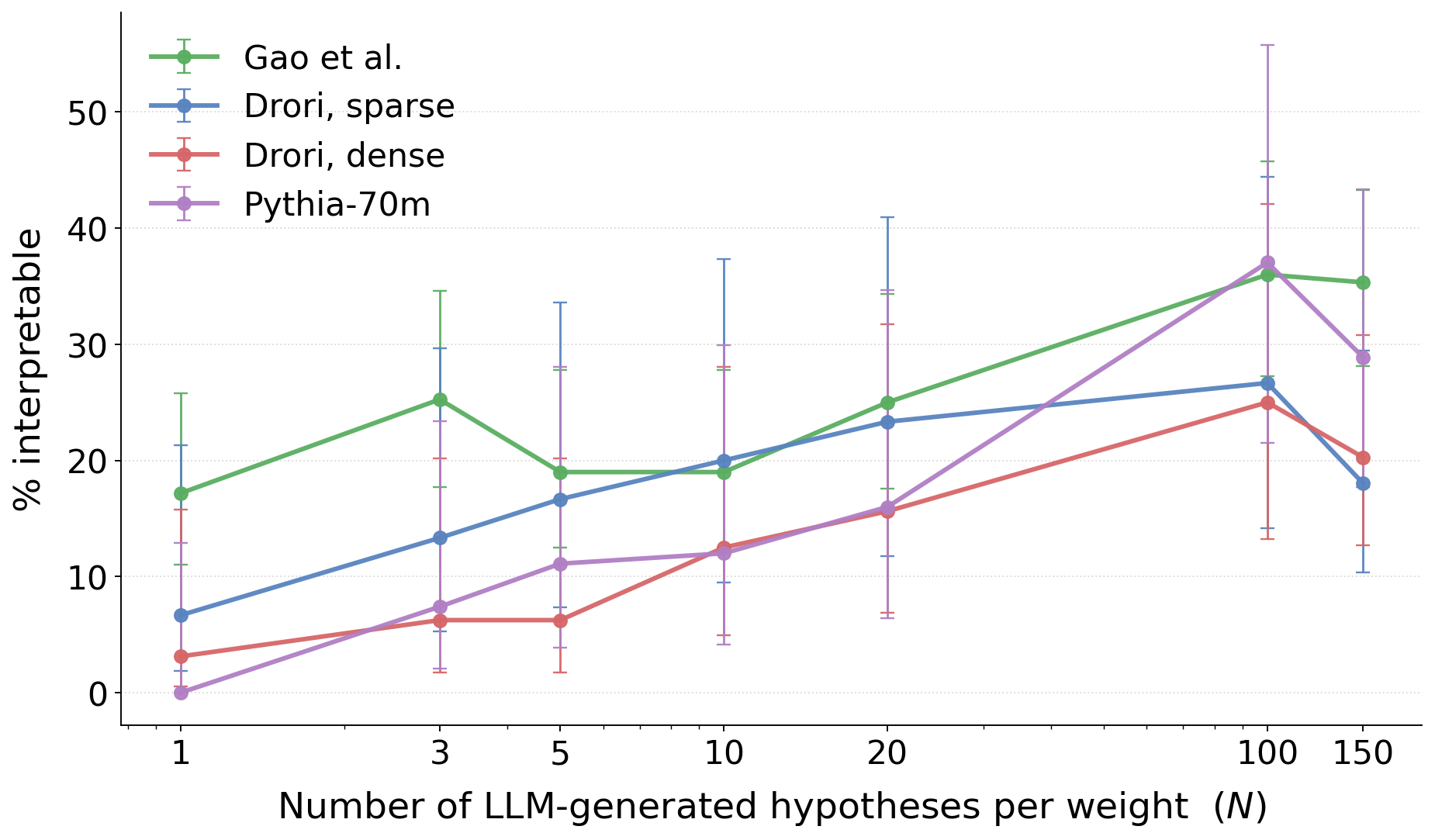}
    \caption{Interpretable rate as a function of the candidate budget $N$, with $95\%$ Wilson CIs. Single-call $N$ from $1$ to $150$.}
    \label{fig:n_sweep}
\end{figure}

A single LLM call is too noisy to call a weight interpretable. Pooling $N$ candidates per weight and selecting the highest-scoring one tightens rates substantially. Sweeping $N \in \{1, 3, 5, 10, 20, 100, 150\}$ (Figure~\ref{fig:n_sweep}) shows interpretability rates rising monotonically with $N$ and saturating before $N = 150$ for the sparse models. We use $N = 100$ as the default because the marginal gain per additional candidate flattens past that point and per-weight LLM cost scales linearly. The cross-model ordering is preserved at every $N$, so the result is not an artifact of the budget we chose.

\subsection{Sensitivity to the auto-interp LLM}
\label{app:llm_comparison}

To check that the result is not specific to one auto-interp LLM, we re-run the pipeline with Claude Sonnet 4.5, GPT-5, and GPT-4o on the same sampled weights.

\begin{figure}[t]
    \centering
    \includegraphics[width=\textwidth]{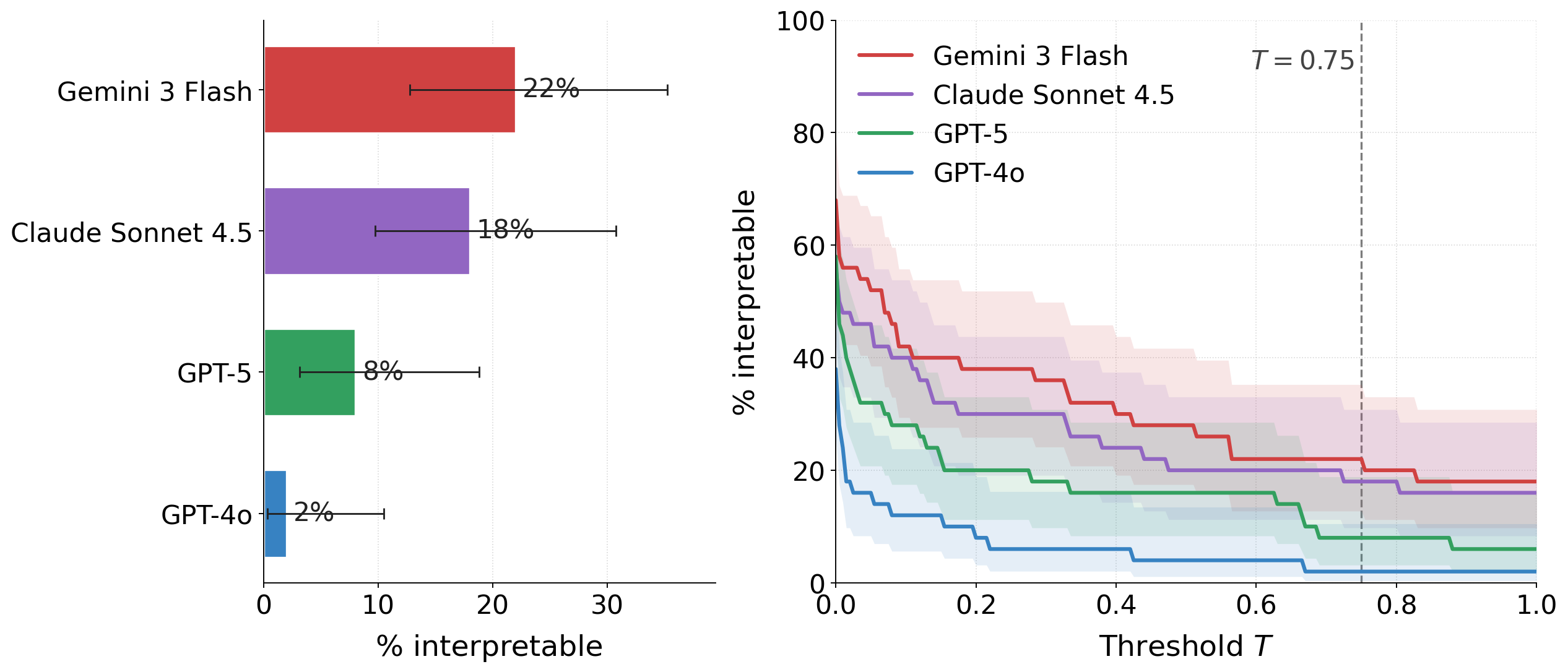}
    \caption{Interpretable rate by auto-interp LLM. \textbf{Left:} fraction of sampled weights interpretable at $T = 0.75$, with $95\%$ Wilson CIs, on the same weights evaluated by four different LLMs. \textbf{Right:} interpretable rate as a function of threshold $T$.}
    \label{fig:llms}
\end{figure}

Figure~\ref{fig:llms} reports the interpretable rate at the canonical threshold $T = 0.75$ for each (model, LLM) pair, and the rate-vs-$T$ curves at $N = 5$. The shape of the curve is consistent across LLMs. Absolute rates differ by up to $\sim$30 percentage points, with stronger LLMs higher, but the cross-model ordering (sparse > dense same-architecture > Pythia) is preserved at every $T$.

\subsection{Digit-detection neuron: full decomposition}
\label{app:digit_neuron}

\paragraph{Per-weight tokens.} Table~\ref{tab:weight_vs_channel_obs} lists, for each of the neuron's nine nonzero input weights, the tokens at the weight's highest-KL positions on the CodeParrot corpus.
\begin{table}[h] \centering \small \caption{For each input weight of the digit-detection neuron (\texttt{neuron 1863}), representative top tokens at the weight's highest-KL positions on the CodeParrot corpus. Tokens are decoded from the tinypython\_2k tokenizer. Rows ordered by max KL.} \label{tab:weight_vs_channel_obs} \begin{tabular}{r r l} \toprule ch ($i$) & max KL & weight top tokens \\ \midrule $464$ & $26.8$ & \texttt{100, 200, 300, 111} \\ $71$  & $22.7$ & \texttt{64, 50, 59, 60} \\ $395$ & $17.1$ & \texttt{28, 08, 27, 35} \\ $799$ & $16.1$ & \texttt{32, 56, 41, 46} \\ $578$ & $13.7$ & \texttt{16, 12, 13, 17} \\ $131$ & $8.0$  & \texttt{4, 2, 3, e} \\ $678$ & $6.0$  & \texttt{7, 8} \\ $87$  & $1.9$  & \texttt{90, 89, 95, f} \\ $558$ & $1.4$  & \texttt{6, 20, 8, 67} \\ \bottomrule \end{tabular} \end{table}

\subsection{String-closing circuit: context-dependent firing}
\label{app:case_quotes}

A weight participating in the quote-closure circuit of \citet{gao2025} suppresses the model's tendency to predict a closing quote at positions that are still inside an open string. This weight's importance depends on context and cannot be inferred from the focus token alone: when we constrain the predicate language to token-local tests (only the focus token, no look-back), the best predicate the LLM finds reaches an interpretability score of $-0.98$, because whether a \texttt{"} token opens or closes a string is undecidable from the token alone. Allowing a small look-back window (\textsc{tokens}[pos$-k$] for $k \leq 8$) takes the score to $+0.83$, well above the interpretability threshold. The weight is interpretable, but only under a predicate language rich enough to look back; the same weight under the token-local language would be filed under ``not interpretable,'' and the circuit would look incomplete from the weights up. We use this finding as the empirical justification for the $\pm 8$-token context window in §\ref{sec:candidates}.

\begin{figure}[t]
    \centering
    \includegraphics[width=\linewidth]{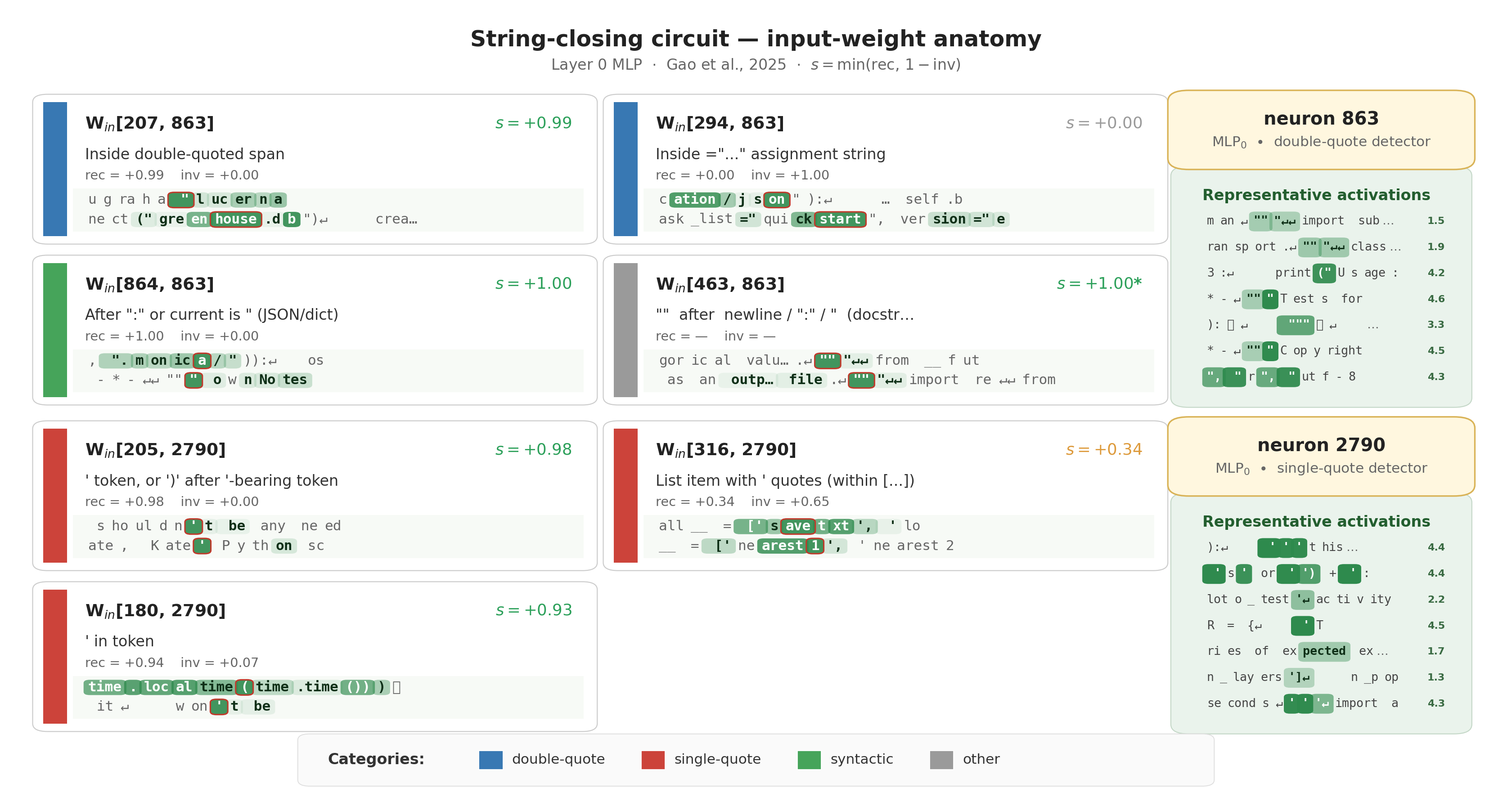}
    \caption{Quote-detector neurons from the string-closing circuit of \citet{gao2025}. \textbf{Neuron 863} (top band) fires on double quotes; \textbf{neuron 2790} (bottom band) fires on single quotes. Each card shows one nonzero input weight: the predicate, the interpretability score $s = \min(\mathrm{rec},\,1-\mathrm{inv})$, the recovery and inverse values, and two top-KL example contexts (focus token outlined in red, pill saturation graded by per-token KL within the row). The right panel of each band shows representative activating contexts of the neuron, randomly sampled from the top-90\% cumulative-activation band on CodeParrot. \\ \textit{Note: $*$ on the score marks weights whose $|\Delta\mathrm{CE}|$ is at the noise floor.}}
    \label{fig:case-quotes}
\end{figure}

Figure~\ref{fig:case-quotes} expands the example to the two quote-detector neurons identified by \citet{gao2025} --- neuron 863 (double quotes) and neuron 2790 (single quotes) --- with the assigned predicate, interpretability score $s = \min(\mathrm{rec},\,1-\mathrm{inv})$, and two top-KL example contexts for each of their nonzero input weights, alongside a panel of representative activations of each neuron drawn from the top-90\% cumulative-activation band.

\subsection{Extended case study: a Drori-sparse speech-verb neuron}
\label{app:case_drori_speech}

This appendix complements the case studies of §\ref{sec:case-studies} with a fully worked example on the Drori sparse model: layer-1 MLP pre-GELU neuron 2461. We show the activation-level view first, then decompose it weight-by-weight, and end by stating what the weight view adds over the activation view.

\begin{figure}[t]
    \centering
    \includegraphics[width=\textwidth]{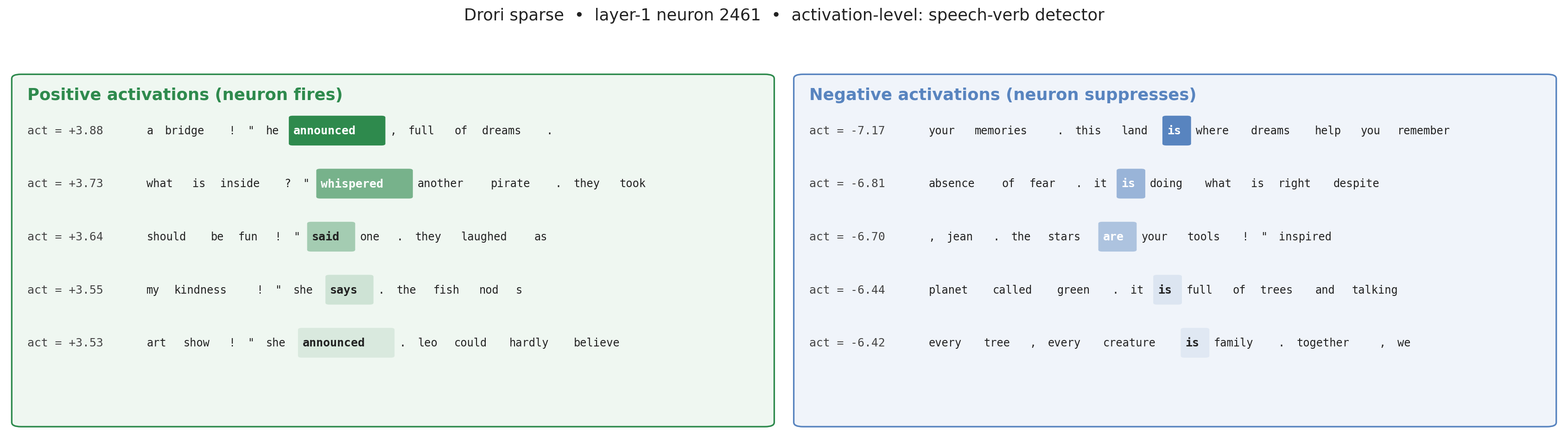}
    \caption{Activation-level view of Drori-sparse layer-1 neuron 2461. \textbf{Left:} five contexts where the pre-activation is most positive (random sample from the top-95\% positive band). The neuron fires on speech-act verbs. \textbf{Right:} five contexts where it is most negative; the neuron is pushed down on copulas (\texttt{is}, \texttt{are}). Pill saturation tracks $|\mathrm{activation}|$ within each panel.}
    \label{fig:case-drori-speech-neuron}
\end{figure}

\paragraph{Activation-level view.} On the SimpleStories corpus, the pre-GELU activation of neuron 2461 has mean $-2.0$, std $1.1$, and a strong skew: the top positive activations sit between $+3.5$ and $+3.9$, and the top negative activations between $-7.2$ and $-6.4$. Figure~\ref{fig:case-drori-speech-neuron} shows five context snippets per direction, randomly sampled from the top-95\% activation band; pill saturation tracks magnitude within each panel. Positive activations (left) are concentrated on speech verbs --- \textit{announced, whispered, said, urged, says} --- in their natural narrative contexts (\textit{``\dots,''~she~announced}, \textit{``\dots''~he~whispered}). Negative activations (right) are dominated by the copulas \texttt{is} and \texttt{are}. At the activation level, this is a \emph{speech-verb detector}: it fires up on speech-act verbs and is pushed strongly down on stative-grammar tokens, in line with what neuron-level autointerp would label.

\begin{figure}[t]
    \centering
    \includegraphics[width=\textwidth]{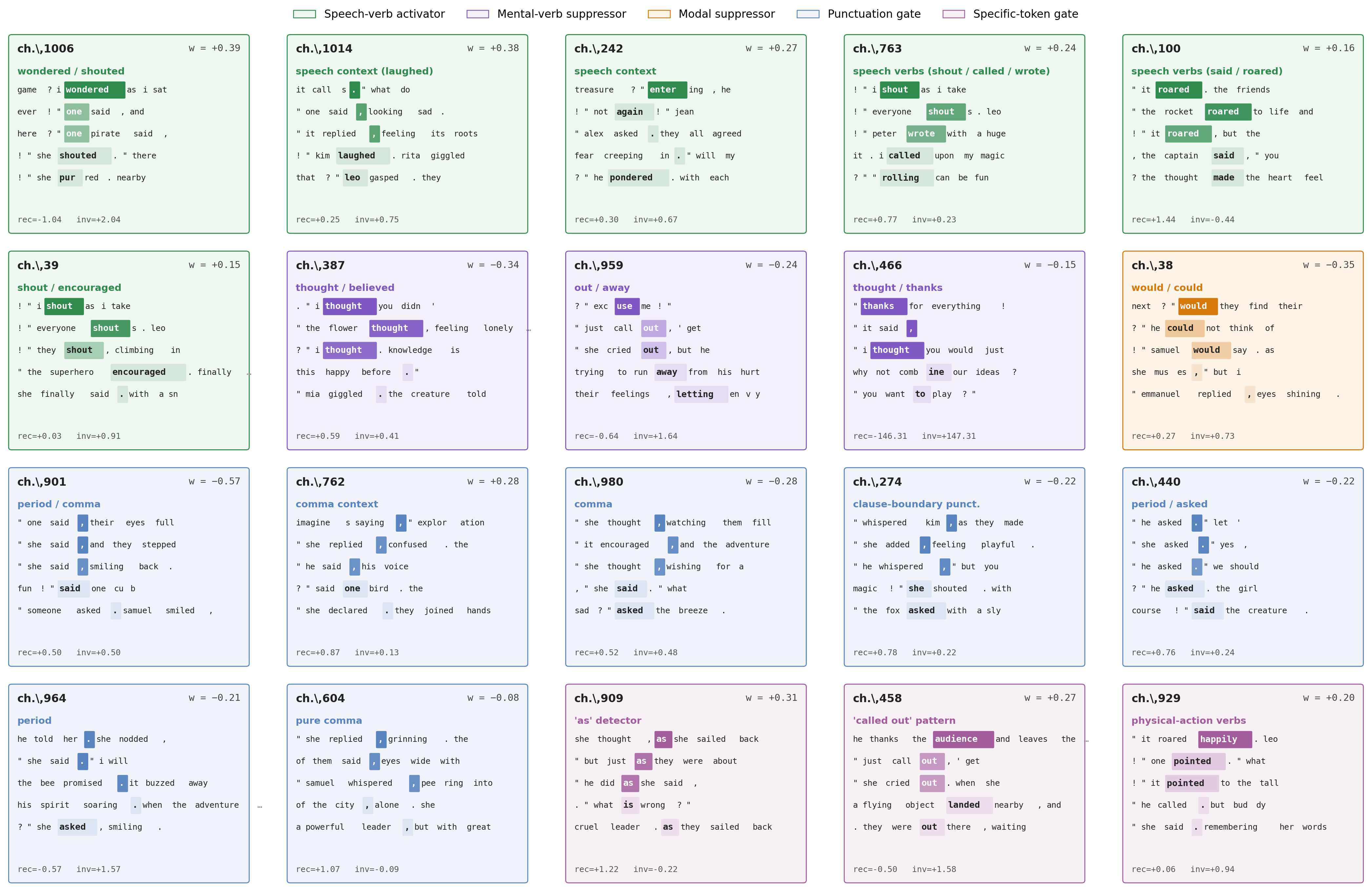}
    \caption{Weight-level decomposition of Drori-sparse layer-1 neuron 2461. The neuron has $86$ nonzero input weights; the figure shows the $20$ strongest by max-KL on a $5{,}000$-sequence SimpleStories sample, grouped by the functional role assigned by their predicate. Five context snippets per card are sampled from the top-95\% cumulative-KL band of that weight; pill saturation tracks the row's KL within the card.}
    \label{fig:case-drori-speech}
\end{figure}

\paragraph{Weight-level view.} The neuron has $86$ nonzero input weights. We rank them by max-KL on a $5{,}000$-sequence SimpleStories sample and profile the top $20$, applying the conditional-zero scoring of §\ref{sec:scoring} to each individually; the remaining $66$ weights have ablation effects below the noise floor of our metric on this corpus and are not analyzed here. Figure~\ref{fig:case-drori-speech} shows all 20 cards, grouped by the functional role assigned by their predicate. Five clusters emerge:

\begin{itemize}
    \item \textbf{Speech-verb activators} (positive weights): six channels whose predicates fire on speech verbs --- \textit{said, roared, whispered, called, wondered, shouted}. These are the weights that drive the neuron up on the left panel of Figure~\ref{fig:case-drori-speech-neuron}.
    \item \textbf{Mental-verb suppressors} (negative weights): three channels whose predicates fire on \textit{thought, believed, understood, thanks}. These actively pull the neuron down on a near-neighbour class of verbs that would otherwise share representational support with speech verbs.
    \item \textbf{Modal-verb suppressor}: one channel firing on \textit{would}, \textit{could}.
    \item \textbf{Punctuation gates}: seven channels whose predicates fire on clause-boundary commas and periods. These contribute to the neuron's negative tail and explain a large share of the \texttt{is}/\texttt{are} suppression observed in the activation view: copulas in this corpus tend to sit one or two tokens after a clause-boundary punctuation mark.
    \item \textbf{Specific-token gates}: three channels with narrower predicates --- a dedicated \textit{`as'} detector, a \textit{`called out'} pattern, and a physical-action verb gate.
\end{itemize}

Each card in Figure~\ref{fig:case-drori-speech} reports recovery, inverse, and the fraction of corpus positions where the predicate fires; the strongest cards (\textit{e.g.,} ch.\,909 \textit{`as'} detector with $\mathrm{rec}=1.22$, $\mathrm{inv}=-0.22$) sit comfortably above the $T=0.75$ interpretability bar. Weaker cards (\textit{e.g.,} the modal suppressor on ch.\,38 with $\mathrm{rec}=0.27$, $\mathrm{inv}=0.73$) do not pass the bar individually, but their predicates remain consistent with the role inferred from neighbouring weights in the same group.

\paragraph{What the weight view adds.} At the activation level neuron 2461 reads as a single ``speech-verb detector.'' At the weight level the same neuron implements \emph{multiple} parallel gating decisions: it activates speech verbs, suppresses near-neighbour verb classes (mental verbs, modals) that share representational support with speech verbs, and routes through clause-boundary punctuation patterns that account for its strongest negative activations. These contributions cancel and add into the single scalar that activation-level analyses report, so they are not separable from the activation summary alone --- only from the per-weight decomposition.

\newpage
\clearpage